%% file: main.tex
\documentclass[10pt,twocolumn,letterpaper]{article}

\usepackage[pagenumbers]{cvpr} 
\input{preamble}
\definecolor{cvprblue}{rgb}{0.21,0.49,0.74}
\usepackage[pagebackref,breaklinks,colorlinks,allcolors=cvprblue]{hyperref}
\usepackage{graphicx}   
\usepackage{booktabs}   
\usepackage{stfloats}
\usepackage{multirow}
\usepackage{float}
\usepackage{stfloats}
\usepackage{cuted}
\def\paperID{9161} 
\def\confName{CVPR}
\def\confYear{2026}

\title{Shape and Texture Recognition in Large Vision-Language Models}

\author{Sagi Eppel\textsuperscript{1}, Mor Bismut\textsuperscript{2}, Alona Strugatski\textsuperscript{1}}

\begin{document}
\maketitle

\let\thefootnote\relax\footnotetext{
\textsuperscript{1} AI Hub, Weizmann Institute 
\textsuperscript{2} Cognitive Science Open University, 
\textsuperscript{} sagieppel@gmail.com. 
The LAS\&T dataset is available at:  \href{https://sites.google.com/view/lastdataset}{\textbf{Main Website}} and \href{https://zenodo.org/records/15453634}{Zenodo}.
}

\input{sec/0_abstract}

\input{sec/1_intro}
\input{sec/1_5_related_work}
\input{sec/2_dataset}

\input{sec/tables}

\input{sec/3_Results_shapes}

\input{sec/4_results_textures}

\input{sec/5_conclusion}

{
    \small
    \bibliographystyle{ieeenat_fullname}
    \bibliography{main}
}

\input{sec/X_suppl}
\end{document}

%% file: sec/0_abstract.tex
\begin{abstract}
\begin{figure*}[b!]
  \centering
 
   \includegraphics[width=1\textwidth]{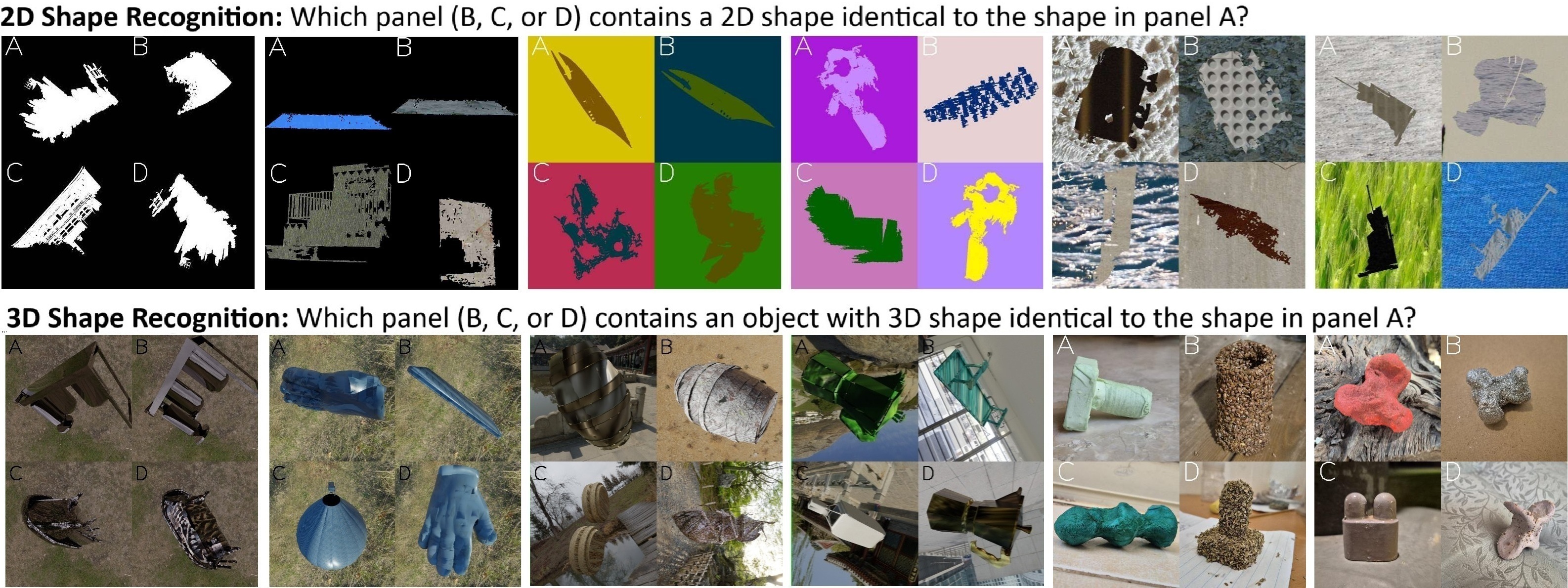}
  \vspace{-1.5em}
  \caption{Testing vision language models' (VLM) ability to recognize shapes and textures using the Large Shape and Textures Dataset (LAS\&T). The models are asked to identify the same shape in different variations, which include different orientations, textures/colors/materials, and environments. The dataset is composed of both simulated and real-world images.
}
  \label{fig:overview}
\end{figure*}
Shapes and textures are the basic building blocks of visual perception. The ability to identify shapes regardless of orientation, texture, or context, and to recognize textures and materials independently of their associated objects, is essential for a general visual understanding of the world. This work introduces the Large Shapes and Textures dataset (LAS\&T), a giant collection of highly diverse shapes and textures, created by unsupervised extraction of patterns from natural images. This dataset is used to benchmark how effectively leading Large Vision-Language Models (LVLM/VLM) recognize and represent shapes, textures, and materials in 2D and 3D scenes.  For shape recognition, we test the models' ability to match images of identical shapes that differ in orientation, texture, color, or environment. Our results show that the shape-recognition capabilities of LVLMs remain well below human performance, especially when multiple transformations are applied. \newline\newline LVLMs rely predominantly on high-level and semantic features and struggle with abstract shapes lacking class associations. For texture and material recognition, we evaluated the models' ability to identify images with identical textures and materials across different objects and environments. Interestingly, leading LVLMs approach human-level performance in recognizing materials in 3D scenes, yet substantially underperform humans when identifying simpler, more abstract 2D textures and shapes. These results are consistent across a wide range of leading LVLMs (GPT/Gemini/LLama/Qwen) and foundation vision models (DINO/CLIP), exposing major deficiencies in the ability of VLMs to extract low-level visual features. In contrast, humans and simple nets trained directly for these tasks achieve high accuracy. The LAS\&T dataset, featuring over 700,000 images for 2D/3D shape and textures recognition and retrieval, is freely available at:  \href{https://sites.google.com/view/lastdataset}{this URL.}

\end{abstract}
\newpage

%% file: sec/1_intro.tex
\begin{figure*}[t]
  \centering
   \includegraphics[width=1\textwidth]{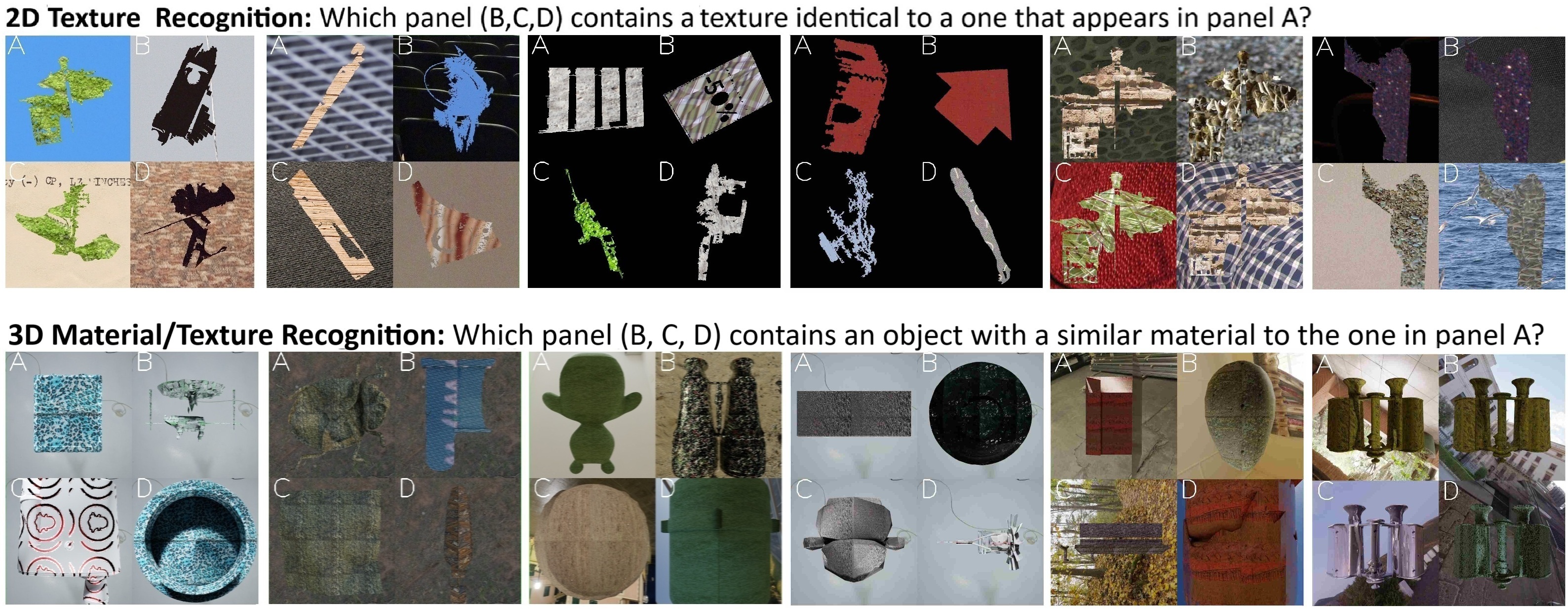}
  \vspace{-1.5em}
  \caption{Samples from LAS\&T dataset for 2D textures and 3D materials recognition and retrieval tests. The model is asked to identify which of the image panels (B,C, or D) contains the same texture/material as the one in panel A. The object/shape on which the material/texture appears, as well as the background and illumination, can change between images.}
  \label{fig:Texture_tests}
\end{figure*}

\section{Introduction}
\label{sec:introduction}
Shapes and textures are the fundamental building blocks of visual perception. The ability to identify and match shapes is a trivial task for humans and is one of the most basic cognitive skill related to spatial perception and object recognition\cite{schwartz1979visual,landau1988importance,attneave1956quantitative,imai1994children}. Similarly, the ability to visually identify and match textures is trivial for humans and is essential for material recognition and understanding of the physical world\cite{landy2004visual,adelson2001seeing,julesz1965texture}. Therefore, it can be expected that any intelligent system that acquires general visual understanding will master both shape and texture recognition. Large vision language models (LVLM) have emerged as the leading computer vision approaches for achieving a general visual understanding of the world\cite{alayrac2022flamingo,li2023blip,chen2025janus,meta2024llama3,bai2025qwen2,achiam2023gpt,team2024Gemini,fu2024blink}. These models are capable of analyzing images and answering complex questions regarding the context of the image in almost any domain and topic\cite{lu2023mathvista,lee2024vhelm,xu2024lvlm}, but at the same time can fail in seemingly simple tasks\cite{rudman2025forgotten,kamoi2024visonlyqa,hemmat2024hidden,eppel2024large}, which brings to question what they actually “see” in the image. 
Most LVLM benchmarks concentrate on higher level tasks like: scene understanding, scientific and logical reasoning, or linking visual inputs to semantic concepts (e.g., scene description, shape classification)\cite{rudman2025forgotten,kamoi2024visonlyqa,lu2023mathvista,lee2024vhelm,xu2024lvlm,hemmat2024hidden,chollet2024arc,zhao2022vl}. In contrast, fundamental perceptual skills such as simple shape or texture matching remain largely unexplored in this domain.
The goal of this work is to test whether leading VLMs like GPT, Claude, Gemini, LLAMA, Qwen, and foundation vision models (CLIP, DINO) have a basic understanding of 2D and 3D shapes as well as textures and materials. Mainly, can the models identify the same shape or object in different settings, even when made of different materials and textures or when viewed with different orientations and environments (\cref{fig:overview}).  Similarly, can the model identify the same textures and materials even when they appear on different objects and environments?  These tasks are trivial for humans\cite{schwartz1979visual,landau1988importance,attneave1956quantitative,imai1994children,landy2004visual,adelson2001seeing,julesz1965texture}  and can provide valuable insight into how models perceive and represent the world. Cognitive and vision research seeks perception blind spots, where humans can miss the basic aspect of a scene (gorilla in the room\cite{chabris2011invisible}), as a way to explore the underlying principles of visual perception\cite{leong2024optical,robinson2013psychology,adelson1993perceptual,coren2020seeing}. Similarly, identifying fundamental gaps in models' performance can help us uncover their internal representation of visual information. \newline
\textbf{Overview on main results: }
The results of these tests show that while all models show some level of shape understanding, they still trail far beyond humans on this task. Transforming aspects of the shape, such as orientation, textures, or background, can lead to a sharp reduction in matching accuracy despite having a minor or no effect on humans. It seems that the models rely on various descriptive features of the shape (ragged, porous, bulky…) and association to semantic classes (cat, house..) and use these features to represent and match shapes. However, in contrast to humans, changing the shape orientation, texture, and even background often changes the descriptive features the model assigns to the same shape. In addition, these descriptive features are often not unique enough to distinguish between different shapes.  If the shape corresponds to an object with a clear semantic class (bird, car…), the models perform far better compared to more abstract shapes that have more low-level distinctive features. This suggests that, unlike humans\cite{imai1994children}, VLM visual representation prioritizes the high-level semantic and class-based concepts tied to the specific shape, and has limited access to the raw shape properties. Another interesting result is that the models seem to perform better in 3D physically realistic scenes compared to the much easier but more abstract 2D scenes. Indicating that task complexity alone does not determine performance; familiarity and the level of abstraction also play a significant role. These results seem to be persistent for both shapes and textures and across a wide range of top VLM and foundation vision models, indicating that despite their ability to understand complex scenes, these models fail to learn some of the most basic aspects of the physical world, and have limited capacity to represent low-level visual features. Interestingly, much simpler nets trained, from scratch, directly on these tasks achieved high precision, suggesting that the training methods and data are the main barrier. 

%% file: sec/1_5_related_work.tex
\section{Related Works}
\label{sec:related_work} 
\textbf{Vision Language Models and Benchmarking:}
Large language models (LLMs) like GPT, Claude, Gemini, Qwen, and DeepSeek have emerged as leading approaches for artificial general intelligence\cite{meta2024llama3,bai2025qwen2,achiam2023gpt,team2024Gemini,guo2025deepseek,liu2024deepseek}. These models have been extended to process images through Vision Language Models (VLMs), also called Large Multimodal Models (LMMs) or Large Vision Language Models (LVLMs)\cite{alayrac2022flamingo,li2023blip,chen2025janus,fu2024blink,lu2024deepseek}. LVLMs map images and text into a unified embedding space and, through training on image captioning and visual question answering (VQA), can answer complex questions across diverse domains\cite{lu2023mathvista,lee2024vhelm,xu2024lvlm}. However, despite their success on high-level tasks like scene understanding, relation prediction, and visual reasoning\cite{chollet2024arc,li2025benchmark,zhang2024vision}, these models can fail at simpler tasks\cite{rudman2025forgotten,kamoi2024visonlyqa,hemmat2024hidden}. While numerous VQA benchmarks evaluate LVLMs on complex conceptual understanding, the far more fundamental capabilities of shape and texture retrieval remain unexplored.\newline
\textbf{Shape and Texture Recognition and Retrieval:}
Shape retrieval involves identifying the exact same shape across different orientations, textures, or environments, distinct from object recognition which assigns semantic classes\cite{shilane2004princeton,kliot1998invariant,cao2022unsupervised,he2018triplet,boyer2011shrec,fu2020hard,lin2021single,chen2003visual,hemmat2024hidden,lin2014microsoft}. Similarly, texture and material retrieval focuses on matching identical materials in varying settings, differing from material classification tasks\cite{drehwald2023one,cimpoi2014describing,lagunas2019similarity,upchurch2022dense,bell2015material}. While specialized models have been developed for these tasks, LVLMs have not been evaluated in this domain.
Existing datasets for 2D shapes and textures rely heavily on manually annotated or synthetic assets limited to small class sets that fail to capture real-world complexity\cite{lin2014microsoft,shilane2004princeton}. Single image-based 3D shape retrieval (IBSR) benchmarks use either synthetic CGI data with limited object classes and backgrounds\cite{fu2020hard,lin2021single}, or real-world images restricted to single categories (predominantly cars)\cite{yang2015large,fu2020hard,sun2018pix3d}.

%% file: sec/2_dataset.tex
\begin{figure*}
  \centering
  \includegraphics[width=1\textwidth]{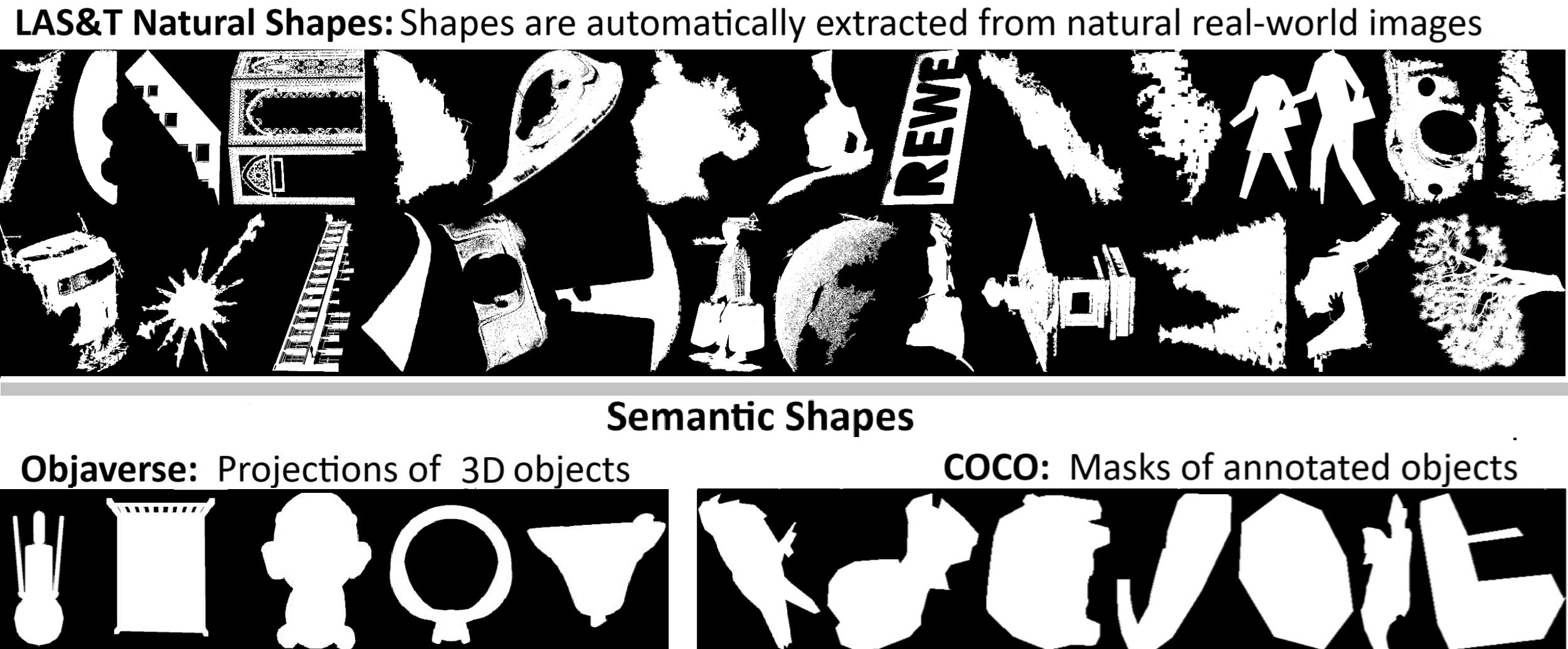}
  \caption{\textbf{Top: LAS\&T dataset, Natural 2D shapes} that were extracted from natural images using an unsupervised
approach (\cref{sec:shape_Extraction,fig:shape_extraction}). \textbf{Bottom: Semantic shapes} corresponding to classes of objects from the
COCO dataset (manual annotations) and 2D projections/silhouettes of 3D objects from the Objaverse
dataset. It can be seen that the diversity and complexity of these semantic shapes are significantly
lower compared to the LAS\&T shapes.
}
  \label{fig:sample_shapes}
\end{figure*}

\section{The Large  Shape \& Textures Dataset (LAS\&T)}
\begin{figure*}
  \centering
   \includegraphics[width=1\textwidth]{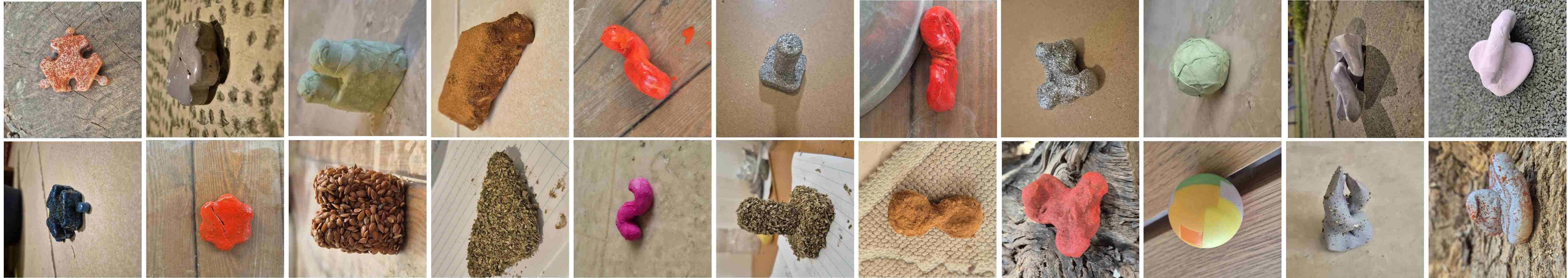}
  \vspace{-1.5em}
  \caption{Samples from the natural images 3D shape retrieval dataset. Each column shows real-world photos of the same 3D shape, coated with different materials and captured under different environments and orientations.}
  \label{fig:real3d}
\end{figure*}

We introduce the Large Shape and Texture dataset (LAS\&T), a large-scale, highly diverse dataset for evaluating shape and texture recognition and retrieval (\cref{fig:overview}).  The dataset is designed to systematically disentangle key visual factors, including 2D/3D, rotation, environment, semantic/natural shapes, textures, and materials, thereby enabling fine-grained analysis of how each element influences visual perception. Evaluation involves tasks where models are asked to identify the same shape under varying textures, orientations, backgrounds, or the same texture/material across different shapes and environments (\cref{fig:overview,fig:Texture_tests}). Another goal of the dataset is to substantially increase the diversity and scale of data available for training and evaluation. Existing datasets either rely on manual annotation, restricting their scale and geometric complexity due to human labor constraints \cite{kirillov2023segment,lin2014microsoft}, or on synthetic data, which, though abundant, are constrained by procedural generation rules and fail to capture real-world complexity \cite{attneave1956quantitative,raistrick2023infinite,nikolenko2021synthetic}. To overcome these limitations, LAS\&T employs unsupervised extraction of shapes directly from natural images (\cref{sec:shape_Extraction}). This yields data constrained only by the variety of available natural images \cite{kirillov2023segment,lin2014microsoft,kuznetsova2020open}.  For 3D shapes, we additionally constructed a general real-world benchmark of 130 photographs, each capturing a 3D shape  coated with different materials in different environments and viewpoints (\cref{sec:ReaImagesBenchmarkcreation,fig:real3d}). This is the first general real-image benchmark for single-image 3D shape retrieval, testing recognition independently of material, pose, or context. LAS\&T comprises four components: 2D Shapes, 3D Shapes, 2D Textures, and 3D Materials, each divided into multiple subsets. As far as we know, each of these sets is the most diverse in its field in terms of different shapes, classes, materials, textures, and environments, often by scale of magnitude (many of the specific tasks in LAS\&T are not covered by any dataset we are aware of). The dataset is  available under a CC0 license.

\begin{figure}
  \centering
   \includegraphics[width=0.46\textwidth]{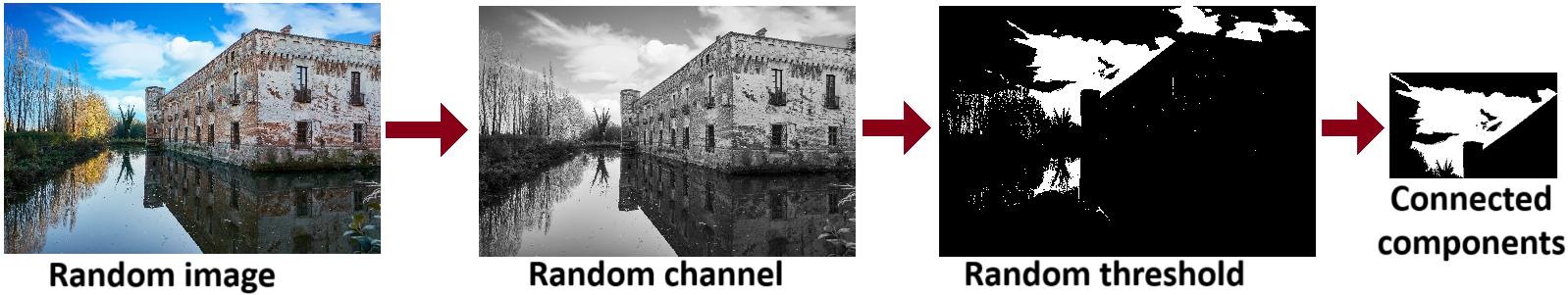}
  \caption{Procedure for automatic shape extraction for the LAS\&T dataset:
a) Randomly select an image.
b) Randomly choose one channel (R, G, B, H, S, or V).
c) Apply a random threshold to binarize the selected channel.
d) Choose a connected component cluster that exceeds the minimum size and thickness criteria and does not touch image boundaries. This can be viewed as a random sampling of natural patterns from images. Examples of extracted shapes are shown in \cref{fig:sample_shapes}.}
  \label{fig:shape_extraction}
\end{figure}

\section{Dataset preparation}
\label{sec:dataset_preperation}
The LAS\&T dataset is designed to evaluate visual recognition of shapes and textures by systematically isolating the effects of various factors, including 3D/2D orientation, backgrounds, environments, semantic versus natural shapes, and textures/materials. Each factor is independently or jointly varied, and a Vision-Language Model (VLM) is tasked with identifying the same shape or texture after transformation. This enables precise mapping of how each element influences recognition, yielding insights beyond those available from existing datasets. In the 2D case, shapes are represented as binary masks of objects or natural image contours (\cref{fig:sample_shapes,sec:shape_Extraction}). Textures are real-world images of uniform textures taken from the  VasTextures dataset\cite{eppel2024vastextures}. Shapes are positioned in random orientations, and both shape and background regions are either filled with a specific texture or kept uniform (black, white, or solid color) (\cref{fig:overview,fig:Texture_tests} top). For 3D scenes, we use the Blender 3D engine to simulate realistic light, matter interactions. Objects are sampled from ObjaVerse\cite{deitke2023objaverse}, covering roughly 1k  different categories and 100k unique 3D models, and rendered with physically based materials (PBRs) from VasTextures (about 100k PBRs). Illumination is provided by 500 distinct High Dynamic Range panoramic Images (HDRI) from the PolyHaven repository\cite{polyhaven_hdri}. Each object is rendered with a selected PBR material and a random HDRI environment. The 3D setup presents far greater challenges than the 2D one, as rotations introduce occlusions, and PBR materials capture realistic reflections, refractions, and complex illumination effects.

\subsection{Natural Shape Extraction and Pattern Infusion}
\label{sec:shape_Extraction}
While 2D shapes can be sourced from segmentation datasets like COCO\cite{lin2014microsoft} or SAM\cite{kirillov2023segment}, which contain millions of human-annotated object masks, these masks tend to have simple, smooth boundaries due to annotation constraints and are semantically tied to real-world objects (\cref{fig:sample_shapes}). To evaluate VLMs’ ability to recognize complex, abstract natural shapes, we propose a new method for sampling shapes directly from images. By thresholding a single image channel (e.g., color or intensity) and extracting connected components above a specified size and thickness (\cref{fig:shape_extraction}), we obtain shapes with intricate, natural boundaries derived from real-world patterns which are not bounded by specific object or semantic classes (\cref{fig:sample_shapes}). This simple method can be seen as a noisy sampling of real-world patterns, producing a far richer and more varied shape distribution than  existing datasets (see \cref{fig:sample_shapes}).

\subsection{Real-World Natural Images Benchmark For 3D Shape Recognition}
\label{sec:ReaImagesBenchmarkcreation}
The 3D shape retrieval component of  the LAS\&T dataset simulates physically realistic material surfaces and light interactions. To validate the simulated data, we built a general real-world images benchmark for 3D shapes retrieval,  where each object maintains the same 3D shape across images, while materials, orientations, and surroundings vary (\cref{fig:real3d}). Objects, ranging from everyday items and rocks to abstract cast and clay forms, were molded or coated with different materials using adhesives to apply the coating. Coatings include granular materials (e.g., sand, spice, coal, grass), paints or various of ribbons and wrappings (\cref{fig:real3d}). Images were captured in varied indoor and outdoor scenes under different lighting and viewpoints, producing 130 real photographs, each representing a distinct object-material pair. This constitutes the first real-world benchmark for evaluating 3D shape retrieval invariant to material, orientation, parts and environment. Meaning the first general dataset in which the 3D shape is the only element used for recognition.

%% file: sec/tables.tex
\begin{table*}
\centering
\caption{Results on 2D shape recognition tests (\cref{fig:overview}).Top rows are various properties that are changed between different images of the same shape. \textbf{Shape type} refers to the source of the shape. \textbf{Natural}: Shapes extracted from natural images (LAS\&T, \cref{fig:sample_shapes}). \textbf{Semantic} are shapes that correspond to specific object classes (COCO, \cref{fig:sample_shapes}). \textbf{Background} and \textbf{Shape} refer to the texture/color of the shape, and background. \textbf{Colored} means uniform random color, and \textbf{textured} means random nonuniform texture. \textbf{Rotation} defines whether the shape was randomly rotated.  \textbf{Text to text}: The model asked to describe each shape independently and a second model made a choice using only this text description (Appendix \cref{fig:query_types}). Columns with the highest variability (all elements changed) are \textbf{bolded}, representing the hardest and most representative test.}
\begin{tabular}{cccccccccc}
\toprule
Shape Type & Natural & Semantic & Natural & Semantic & Natural & Natural & Natural & Natural & Natural \\

Background & Colored & Colored & Textured & Textured & Colored & Black & Black & Black & Textured \\
Shape & Colored & Colored & Textured & Textured & Colored & White & White &  Textured & Textured \\
Rotation & V & V & V & V & X & V & X & X & X \\
\midrule
Gemini-1.5pro & 0.90 & 0.95 & \textbf{0.58} & 0.77 & 0.97 & 0.98 & 1.00 & 0.85 & 0.78 \\
Gemini 1.5 pro text to text & 0.84 & 0.92 & \textbf{0.68} & 0.80 & 0.96 & 0.91 & 0.97 & 0.87 & 0.80 \\
\hline
GPT 4O & 0.94 & 0.99 & \textbf{0.79} & 0.91 & 0.99 & 0.97 & 0.99 & 0.95 & 0.89 \\
GPT 4O text to text & 0.90 & 0.96 & \textbf{0.77} & 0.88 & 0.98 & 0.96 & 0.98 & 0.92 & 0.86 \\
\hline
Gemini-2.0-flash & 0.79 & 0.91 & \textbf{0.55} & 0.71 & 0.92 & 0.92 & 0.97 & 0.86 & 0.68 \\
Gemini 2.5 flash & 0.94 & 0.96 & \textbf{0.72} & 0.78 & 0.98 & 0.97 & 0.98 & 0.91 & 0.82 \\
Claude 3.5 Sonnet & 0.89 & 0.94 & \textbf{0.63} & 0.79 & 0.97 & 0.90 & 0.97 & 0.90 & 0.73 \\
Grok-2-vision & 0.93 & 0.97 & \textbf{0.77} & 0.85 & 0.99 & 0.96 & 0.99 & 0.91 & 0.84 \\
Llama 4 Scout 17B & 0.90 & 0.96 & \textbf{0.63} & 0.79 & 0.95 & 0.95 & 0.91 & 0.81 & 0.74 \\
Qwen2 VL 72B & 0.87 & 0.95 & \textbf{0.61} & 0.63 & 0.95 & 0.95 & 0.94 & 0.74 & 0.67 \\
Gemini-2.0-flash-thinking & 0.89 & 0.96 & \textbf{0.66} & 0.73 & 0.97 & 0.99 & 0.98 & 0.89 & 0.79 \\
o4 mini & 0.96 & 0.99 & \textbf{0.83} & 0.92 & 0.99 & 0.99 & 0.98 & 0.96 & 0.94 \\
\hline
DINO V2 Descriptor & 0.94 & 0.96 & \textbf{0.35} & 0.40 & 0.99 & 1 & 1 & 0.81 & 0.42 \\
CLIP Descriptor & 0.76 & 0.83 & \textbf{0.40} & 0.46 & 0.98 & 0.98 & 1.00 & 0.83 & 0.48 \\
DINO V2 Extracted & 0.94 & 0.94 & \textbf{0.63} & 0.62 & 0.98 & 0.97 & 0.97 & 0.88 & 0.65 \\
CLIP Extracted & 0.94 & 0.94 & \textbf{0.80} & 0.77 & 0.95 & 1.00 & 0.98 & 0.85 & 0.73 \\
\hline
Specialize Net  & 1 & 1 & \textbf{1} & 1 & 1 & 1 & 1 & 1 & 1 \\
Humans & 1 & 1 & \textbf{1} & 1 & 1 & 1 & 1 & 1 & 1 \\
\bottomrule
\end{tabular}
\label{table:2dshapes}
\end{table*}

\begin{table}[t]
\centering
\small
\caption{Result for 3D shape recognition (\cref{fig:overview}).  \textbf{V} means the property is changed between scenes. \textbf{X} means the property remains constant. \textbf{Sim} refers to simulated/synthethic images. \textbf{Real} refer to natural images (\cref{sec:ReaImagesBenchmarkcreation,fig:real3d}). Columns with the highest variability (all elements changed) are \textbf{bolded}, representing the hardest and most representative test.}
\begin{tabular}{lccccc}
\toprule
Image Type & Sim & Sim &  Sim & Sim & Real \\
Varied Orientations & V & X & V & V & V \\
Varied Textures & X & V & V & V & V \\
Varied Backgrounds & X & X & X & V & V\\
\midrule
Claude-3.5-sonnet & 0.82 & 0.88 & 0.74 & \textbf{0.69} & \textbf{0.70}\\
Gemini-1.5-pro & 0.87 & 0.93 & 0.83 & \textbf{0.81} & \textbf{0.79}\\
Gemini 2.5 Flash & 0.92 & 0.98 & 0.87 & \textbf{0.80}  & \textbf{0.83}\\
GPT 4o & 0.89 & 0.96 & 0.85 & \textbf{0.82} & \textbf{0.85}\\
Grok 2 & 0.92 & 0.95 & 0.89 & \textbf{0.82} & \textbf{0.89}\\
o4 Mini & 0.93 & 0.98 & 0.91 & \textbf{0.88} & \textbf{0.90}\\
Llama 4Scout 17B & 0.81 & 0.90 & 0.80 & \textbf{0.79} & \textbf{0.80}\\
Qwen2 VL 72B & 0.84 & 0.93 & 0.79 & \textbf{0.75} & \textbf{0.70}\\

\hline
DINO V2 Descriptor & 0.88 & 0.96 & 0.83 & \textbf{0.73} & \textbf{0.90}\\
CLIP Descriptor & 0.79 & 0.91 & 0.67 & \textbf{0.55} & \textbf{0.78}\\
DINO V2 Extracted & 0.88 & 0.96 & 0.82 & \textbf{0.83} & \textbf{0.86}\\
CLIP Extracted & 0.83 & 0.90 & 0.84 & \textbf{0.80} & \textbf{0.80}\\
\hline
Specialize Net & 1 & 0.98 & 0.98 & \textbf{0.97} & \textbf{0.94}\\
Human & 0.98 & 0.98 & 0.98 & \textbf{0.97} & \textbf{1}\\
\bottomrule
\end{tabular}
\label{table:3dshapes}
\end{table}


\begin{table}[t]
\centering
\small
\caption{
Result for 2D texture retrieval (\cref{fig:Texture_tests}). The top two rows refer to elements of the image that are changed between different occurrences of the same texture. \textbf{Background} refers to the background of the texture, which can be \textbf{black} or covered with a different random texture (\textbf{txt}). \textbf{Shape} refers to the shape the texture appears on, this can be the \textbf{same} shape for all instances of the texture or \textbf{dif}ferent shape for every instance (\cref{fig:Texture_tests}). Columns with the highest variability (all elements changed) are \textbf{bolded}, representing the hardest and most representative test.}
 
\begin{tabular}{lcccc}
\toprule
Background & txt & txt & Black & Black \\
Shape & dif &  same & dif & same \\
\midrule
Grok-2 & \textbf{0.82} & 0.88 & 0.91 & 0.97 \\
Claude-3.5 sonnet & \textbf{0.64} & 0.72 & 0.82 & 0.94 \\
GPT 4O & \textbf{0.88} & 0.92 & 0.93 & 0.99 \\
GPT 4O-mini & \textbf{0.59} & 0.60 & 0.66 & 0.74 \\
o4 mini  & \textbf{0.88} & 0.91 & 0.97 & 0.99 \\
Qwen2 VL 72B & \textbf{0.60} & 0.66 & 0.69 & 0.83 \\
Llama4 Scout 17B & \textbf{0.75} & 0.81 & 0.91 & 0.96 \\
Gemini-2.0-flash & \textbf{0.71} & 0.79 & 0.81 & 0.97 \\
Gemini-2.5-flash  & \textbf{0.82} & 0.87 & 0.91 & 0.98 \\

\midrule
DINO V2 Desc & \textbf{0.28} & 0.51 & 0.51 & 0.80 \\
CLIP Desc & \textbf{0.36} & 0.49 & 0.59 & 0.94 \\
DINO V2  Extract & \textbf{0.51} & 0.55 & 0.59 & 0.82 \\
CLIP  Extract& \textbf{0.78} & 0.82 & 0.89 & 0.98 \\
\midrule
Specialize Net & \textbf{1} & 1 & 1 & 1 \\
Humans & \textbf{0.98} & 0.99 & 0.99 & 0.99 \\
\bottomrule
\end{tabular}
\label{table:2d_textures}
\end{table}

\begin{table}[t]
\centering
\small
\caption{
Result for 3D material recognition (\cref{fig:Texture_tests}). The top two rows refer to elements of the image that are changed between different occurrences of the same material. \textbf{Background} refers to the background and illumination. \textbf{Object} refers to the object the material appears on (\cref{fig:Texture_tests}). \textbf{V}=Changed, \textbf{X}=constant. Columns with the highest variability (all elements changed) are \textbf{bolded}, representing the hardest and most representative test.}

\begin{tabular}{lccc}
\toprule
Object & V & X & V \\
Background & X & V & V \\
\midrule
Grok-2 & 0.93 & 0.96 & \textbf{0.87} \\
GPT 4O & 0.96 & 0.95 & \textbf{0.91} \\
Claude-3-5 & 0.68 & 0.77 &  \textbf{0.62} \\
Gemini-1.5-pro & 0.88 & 0.88 &  \textbf{0.74} \\
Gemini-2.0-flash & 0.87 & 0.81 &  \textbf{0.77} \\
Qwen2 VL 72B & 0.76 & 0.85 &  \textbf{0.59} \\
Llama4 Scout 17B & 0.86 & 0.92 &  \textbf{0.71} \\
o4 mini & 0.92 & 0.94 &  \textbf{0.82} \\
Gemini-2.5-Flash & 0.82 & 0.93 &  \textbf{0.66} \\
\midrule
DINO V2 Descriptor & 0.48 & 0.63 & \textbf{0.27} \\
CLIP Descriptor & 0.68 & 0.61 & \textbf{0.42} \\
DINO V2 Extracted & 0.75 & 0.84 & \textbf{0.77} \\
CLIP Extracted & 0.77 & 0.70 & \textbf{0.54} \\
\midrule
Specialize Net & 0.97 & 0.97 & \textbf{0.93} \\
Human & 0.95 & 0.98 & \textbf{0.93} \\
\bottomrule
 
\end{tabular}
\label{table:3d_materials}
\end{table}

%% file: sec/3_Results_shapes.tex
\section{Shape recognition: Results and discussion}
\label{sec:results_shape}
 The results of the shape recognition tests (\cref{fig:overview}) are given in \cref{table:2dshapes,table:3dshapes}. It's clear that all models have understood the task and can recognize both 2D and 3D shapes under various transformations.  This can be seen both from the fact that all models performed way above a random guess, and achieved high accuracy for simpler matching tasks (\cref{table:2dshapes,table:3dshapes}). In addition, when asked to explain their answers (both correct and incorrect), the models refer to the real features of the shape in each panel (For example, responses see Appendix \cref{appendix:2Dshapes_model_response,appendix:3Dshapes_model_response}).  However, all models perform well below humans (\cref{table:2dshapes,table:3dshapes}), the gap is particularly large for 2D and when multiple transformations are applied. For example, when both texture and orientation of the 2D shape were changed (\cref{fig:overview,table:2dshapes}), the matching accuracy of all models was in the range of 40-80\%  compared to perfect 100\% for humans and 33\% for a random guess.\newline
\textbf{What do the models see?}
When asked to explain the answer, the LVLMs usually refers to a set of high-level descriptive features of the shape, like rough/narrow/bulky… or to similarity to a given set of classes (bird, house..). Samples of the explanations given by different models are available in Appendix \cref{appendix:2Dshapes_model_response,appendix:3Dshapes_model_response}.
Errors and mismatches often occur because the model assigns different descriptions to the same shape when it appears in different orientations or textures. Although both descriptions may be valid, they are often inconsistent with each other.
  In addition, mistakes often arise from different shapes sharing the same high-level features (for example, two shapes being bulky, or curved) or from different shapes with the same perceived class (for example, two different shapes resemble a house).  However, it does seem that the top-performing models (GPT, Gemini, Grok) can easily distinguish between the shape, texture, and background and describe each separately.
It should be noted that the verbal explanation the model gives to explain their answer (when asked to explain) strongly correlates with both the shapes in the image and the selected answer, suggesting that the decision process is clearly related to semantic concepts (See Appendix \cref{appendix:2Dshapes_model_response}). At the same time, it seems that the models do not have access to the full shape information and only rely on a limited set of high-level features.\newline
\textbf{Semantic vs natural shapes:}
The source of the shapes seems to have a significant effect on the accuracy of the models. All LVLMS were able to identify and match 2D shapes that directly correspond to silhouettes of objects related to classes with higher accuracy compared to more abstract 2D shapes, despite the latter having far more distinctive features. (\cref{fig:sample_shapes}, \cref{table:2dshapes} semantic vs. natural shapes). This again implies that the way LVLM understands images is by mapping the content into a high-level representation closely related to language concepts; when the shape does not strongly correspond to a unique semantic class, the accuracy drops. Given that LVLMs are usually trained by the image-to-text questions, it is likely that these models are incentivized to map the image into semantic-like concepts. This is in contrast to humans, for whom the vast majority of visual tasks are not related to language and therefore can deal with the shape much more directly\cite{schwartz1979visual,landau1988importance}. This makes abstract visual understanding of VLM very limited compared to humans, who perform almost perfectly on all tests (\cref{table:2dshapes,table:3dshapes} Humans). \newline
\textbf{Is the image verbal description enough?}
\label{text2text}
To further test the hypothesis that the representation of the image in LVLM is strongly related to the semantic/verbal description, we try the following experiment. First, an LVLM was asked to independently describe the shape in each panel without referring to other panels in the image.   Then another LVLM received the generated textual description without the image and was asked to determine which of the shapes match based on their description (For more  details, see Appendix \cref{fig:query_types}). 
The model that uses only the verbal description of each shape as input (referred to as text-to-text in \cref{table:2dshapes}) achieves similar or slightly lower accuracy than the same LVLM that uses the image directly (\cref{table:2dshapes}.text-to-text). This implies that the verbal description is mostly enough to explain the matching accuracy.  Hence, whatever internal representation the models have when dealing with the image, it does not contain significantly more relevant details than the verbal description the model gives to the shape.  Once again, supporting the hypothesis that the internal representation of the images is strongly related to verbal and semantic concepts.\newline
\textbf{What information do foundation vision models extract?} 
The most direct way to examine what models "know" about shapes is by analyzing their internal image representations or embeddings. While access to the embeddings and training data of models like GPT is not available. There are several open-source foundation vision models that give full access to their embedding. OPEN CLIP and DINO V2 are two of the most commonly used open source foundation vision models that generate such embeddings\cite{schuhmann2021laion,oquab2023dinov2}. Accessing these embeddings allows us to probe the extent to which shape information is encoded. A basic test involves checking whether embeddings of the same shape (with variations in texture or orientation) are more similar to each other than to embeddings of different shapes. Using cosine similarity, we compared the embeddings for images of similar and different shapes (See Appendix \cref{fig:clipdino}). As shown in \cref{table:2dshapes,table:3dshapes} (CLIP/DINO descriptors), the matching accuracy is comparable to that of LVLMs, but drops significantly in cases with large visual variations. However, embeddings from these models capture more than just shape. Therefore, embedding similarity does not necessarily reflect shape similarity, especially in complex images. To isolate shape-specific information, we trained an extraction network on the LAS\&T dataset. This net takes a CLIP or DINO embedding as input and produces a new embedding that only represents shape similarity, filtering out unrelated visual information. Hence, the extraction net retrieves whatever shape-related information is present in the original embedding  and discard the rest (See Appendix \cref{fig:clipdino}). The performance of this refined descriptor (\cref{table:2dshapes}.DINO/CLIP extracted) is better in most cases compared to the original descriptor. Still, performance remains well below human-level accuracy. This suggests that CLIP and DINO embeddings do not contain the full information about the shape. Nets trained specially for the task  of extracting shape descriptors from images (Appendix \cref{fig:speciallizenet}) achieve near-perfect shape-matching accuracy (\cref{table:2dshapes,table:3dshapes}.Specialize Net). This implies the limitation lies not in the extraction net or its training, but in the information content of the CLIP/DINO embeddings. 
\newline
\textbf{3D vs 2D  and Difficulty vs Abstraction:}
Another interesting result is that the accuracy of 3D shape matching tests is similar and even higher than that of 2D shape matching tests (\cref{table:2dshapes,table:3dshapes}). This is surprising since 3D shape matching demands an understanding of occlusion perspective and illumination, making it far more complex than 2D. This result seems to imply that the complexity of the task is not the only factor that affects the model's performance. It's likely that the models that were trained on large amounts of real-world 3D scenes\cite{schuhmann2021laion} have gained a better understanding of the more complex 3D objects, but have far less exposure to the simpler but also more abstract 2D shapes.
\newline
\textbf{Shape matching difficulty and specially trained net:}
The relative difficulty of the shape matching test can be estimated by training specialized nets for this task using the LAS\&T dataset. Separate nets were trained to match shapes in 2D and 3D  using images as input. These nets were trained to receive images containing shapes with various variations and generate a descriptor vector that represents  the shape's similarity (Appendix \cref{fig:speciallizenet}). The trained nets easily achieved almost perfect accuracy in matching 2D shapes (\cref{table:2dshapes}.Specialize net) and lower but still high accuracy in 3D shape matching (\cref{table:3dshapes}.Specialize net), confirming that 3D shape matching is objectively harder. These results confirm that the problem the LVLM has with shape recognition is not a result of the objective difficulty of the task, but rather of the way these models were trained. Training details and code, and weights for the LAS\&T trained nets are available in the S.I (Appendix  \cref{fig:clipdino,fig:speciallizenet}). \newline

%% file: sec/4_results_textures.tex
\section{Texture and material recognition: Results and discussion}
\label{sec:ResultsTextureMaterialRecognition}
For material recognition in 3D scenes, the models achieve relatively high accuracy with the highest performing model reaching almost human level, even when both object and environment changed (\cref{table:3d_materials}).  Note that this is a relatively complex task even for humans, since materials can look quite different under different illumination conditions (\cref{fig:Texture_tests}).  Interestingly, for the much easier 2D texture recognition test (\cref{table:2d_textures}), the models performed far below humans, and with accuracy similar to that of the 3D material recognition tests (\cref{table:3d_materials}). Note that 2D texture recognition involves matching the exact same textures only overlayed on different shapes and in different backgrounds (\cref{fig:Texture_tests}), a task that is trivial for humans.  These results again show that the complexity of the task is not necessarily the main factor in the performance of the models. Again, these results might be explained by the fact that vision language models are trained mostly on real-world images and, therefore far more familiar with real materials in different 3D environments than with the simpler but more abstract 2D textures.
Humans achieve high accuracy on all tasks and consistently perform better on 2D texture recognition than on 3D material recognition. Nets trained from scratch on this task using the LAS\&T dataset achieve almost perfect accuracy and higher accuracy in 2D compared to 3D scene (\cref{table:2d_textures,table:3d_materials} specialized net). Confirming that LVLM limitations are a result of the training methods and not the fundamental limitation of the net. \newline
\textbf{What does the model see?}
When asked to explain their answer, the models refer to specific textures, colors, and materials features (metallic, stripes, or rock...), which proves that the model understands both the image and the task (Appendix \cref{appendix:2D_texture model Respons,appendix:3d_materials_response}). For 3D materials, the main cause of the error is having the same material appearing in different illumination or confusing different materials that share a high similarity. We note that the model failure cases for 3D materials are challenging to humans as well (Appendix \cref{appendix:2D_texture model Respons,appendix:3d_materials_response}). For 2D textures (\cref{fig:Texture_tests}), one source error seems to be mixing of properties from the background texture or the shape outline with the properties of the actual texture, implying that in some cases, the ability of the model to separate and focus on the main texture is the challenge (this is again trivial for humans).

%% file: sec/5_conclusion.tex
\section{Conclusion} 
\label{sec:conclusion}
Shape and texture are the fundamental building blocks of visual perception, making their recognition and representation essential for any system seeking to achieve general visual understanding. This work explores how vision-language models (VLMs) recognize and represent these elements using a novel dataset that isolates each visual component's effect. This separation enables us to gain new insights into how leading VLMs represent these concepts and reveals some of their blindspots and limitations. 
Our analysis revealed several interesting findings:\newline
\textbf{Human-Level Gaps:} LVLMs show some understanding of shapes and textures but perform well below human levels, especially in abstract cases.\newline
\textbf{2D vs. 3D Performance:} The models performed similarly or even better on 3D shape and texture recognition compared to 2D. This is surprising given the added complexity of 3D tasks, which involve occlusion, perspective, and lighting. These results suggest that model performance is not strictly determined by task complexity, but also by the distribution of training data, which often emphasizes real-world 3D scenes.\newline
\textbf{Semantic Bias:} VLM representations of shapes rely heavily on semantic descriptions. Models performed substantially better on shapes associated with recognizable object categories than on abstract shapes, despite the latter having far more distinctive visual features. In contrast, human participants showed no such bias, achieving near-perfect accuracy in both conditions. Furthermore, when models were asked to first describe an image verbally and then perform shape matching using only that description, their performance was similar to cases where they directly processed the image. This suggests that VLMs' internal shape representations closely parallel their verbal descriptions rather than capturing raw visual details.
This indicates that LVLMs tend to rely on high-level semantic features and class associations rather than comprehensive low-level shape representations. These findings suggest that current LVLM training paradigms, while effective in real-world object recognition, may not provide the level of visual abstraction required for human-level perception.
The LAS\&T dataset, with its large-scale and highly diverse collection of patterns and images, provides a valuable resource for addressing this limitation, serving as both training data and a benchmarking tool.

%% file: sec/X_suppl.tex
\clearpage
\setcounter{page}{1}
\maketitlesupplementary

\section{Dataset code and weights availability}
The complete LAS\&T dataset, as well as the scripts used to generate the synthetic data  and run the tests, combined with  the code and weights for the nets trained in this work, are freely under the CC0 license, from \href{https://sites.google.com/view/lastdataset/}{LAS\&T dataset web}, \href{https://zenodo.org/records/15453634}{Zendo} and  \href{https://github.com/sagieppel/Shape-and-Texture-recognition-in-large-vision-language-models-}{GitHub}.

\begin{figure*}
  \centering
   \includegraphics[width=1\textwidth]{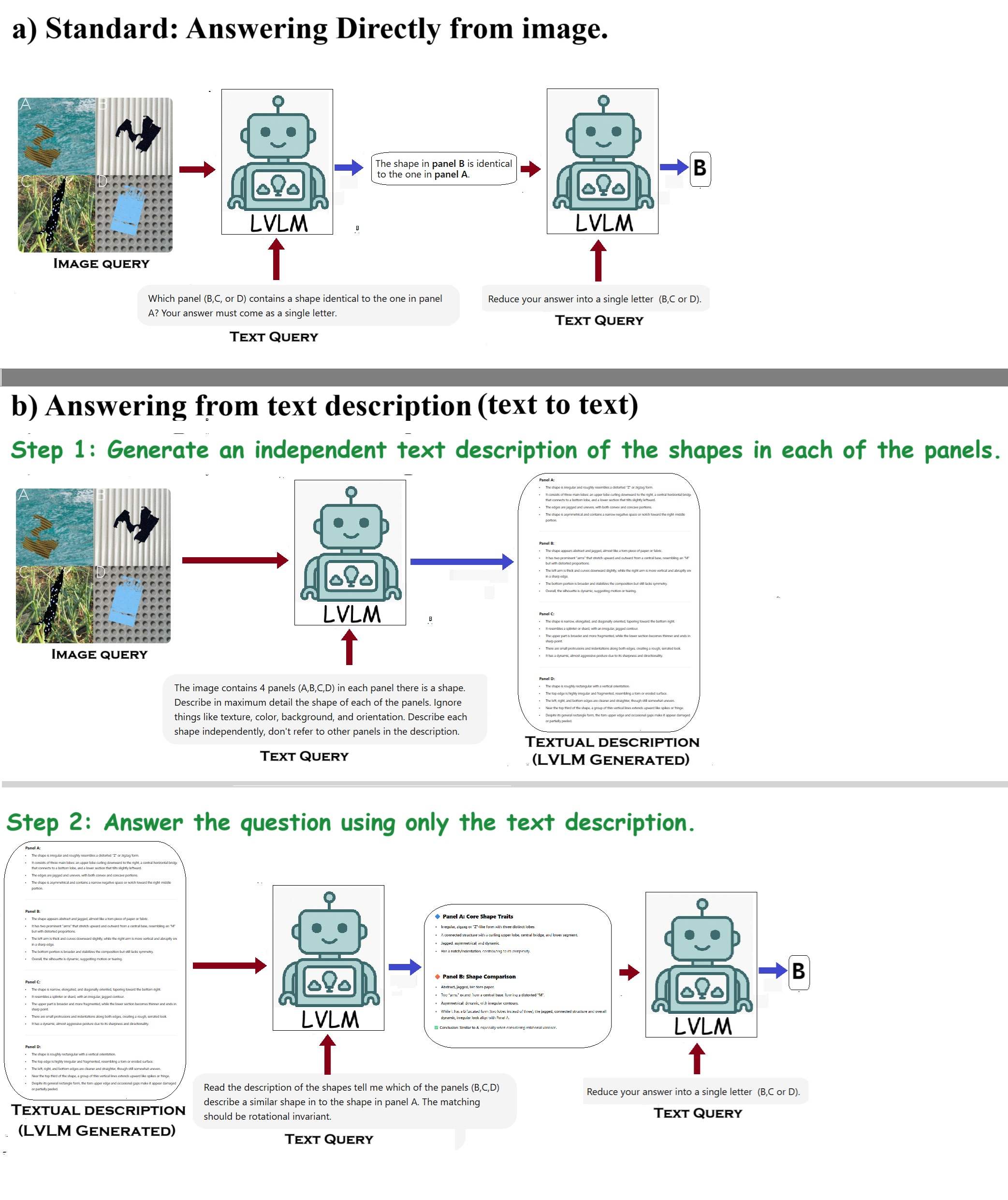}
  \vspace{-1.5em}
  \caption{Format of LVLM automatic testing used to generate the results in \cref{table:2dshapes,table:3dshapes,table:2d_textures,table:3d_materials}.\newline a) Standard testing method: the model receives the image and a text query asking which of the panels (B,C,D) contain a similar shape to the one in panel A. If the answer is not a single letter a second model is used to extract the answer as a single letter. \newline b) Testing based on textual description (\cref{text2text,table:2dshapes} text to text). The model is asked to describe the shape in each panel independently (without referring to other panels). The generated text description is given to a second model with no memory or access to the image. The second model is asked to decide which of the panels (B,C,D) contain a similar shape to the one in panel A. The fact that both methods (a,b) give similar accuracy (\cref{table:2dshapes} text to text) implies that the information the model extracts regarding the shapes is mostly contained in the textual description it gives the shape.
}
  \label{fig:query_types}
\end{figure*}
\section{Testing details}
\label{appendix:testing_details}
\subsection{Choosing text query}
An important aspect of  LVLM testing is how to express the query given to the model. While in theory, any clear description of the task should work, in practice different prompts can lead to up to 8\% difference in accuracy. Interestingly detailed and exact prompts perform worse than short general ones. For example the short prompt: \newline
\textbf{“Which of the panels contains an object with an identical 3D shape to the object in panel A. Your answer must come as a single letter”} \newline
\newline
Give better accuracy than the long detailed prompt like: \newline
\textbf{“Carefully analyze the image. In panel A, there is an object with a specific shape. Your task is
to identify which other panel (B, C, or D) contains an object that\newline
1) Has the exact same 3d shape as the object in panel A.\newline
2) Has a different orientation compared to the object in panel A.\newline
3)Has a different texture compared to the object in panel A.\newline
Respond with ONLY the letter of the panel (B, C, or D) that meets all these criteria.”}
\newline

Specifically, a detailed description of the various possible transformations seems to have negative effects. Asking the models to write their own prompts led to longer, more detailed text, which led to worse results. One explanation for this is that the models understand the task quite well from the short prompt and the longer prompt just consumes more attention. There are few strong indications that all models clearly understood the task. First when asked to explain their choices they clearly give explanations that are consistent with the tasks and the object in each panel (\cref{appendix:2Dshapes_model_response}). Second, the accuracy of the answers while below human is well above random (33\%) and is higher and easier the task is (\cref{table:2d_textures,table:3d_materials,table:2dshapes,table:3dshapes}).
\subsection{Automatic Extraction of the panel from the model answer.}
Testing the models was done automatically with 400 questions per tests using the models APIs (for open source models we used Together.AI). This was done automatically by asking the models to give only the current panel (\cref{fig:query_types}) in the answer.
 However, models sometimes give long answers even when asked to only give the panel letter, and sometimes refuse to answer claiming that no panel contains an identical shape. In these cases, the question was repeated several times as is and then with greater assertiveness. While in case of a long answer, the model was asked to shorten the answer to one letter(\cref{fig:query_types}), we manually verified that this rarely if ever adds error to the original answer. For each task, several prompts were tested, and the best results are given in \cref{table:2d_textures,table:3d_materials,table:2dshapes,table:3dshapes}. 

\begin{figure*}[t]
  \centering
   \includegraphics[width=1\textwidth]{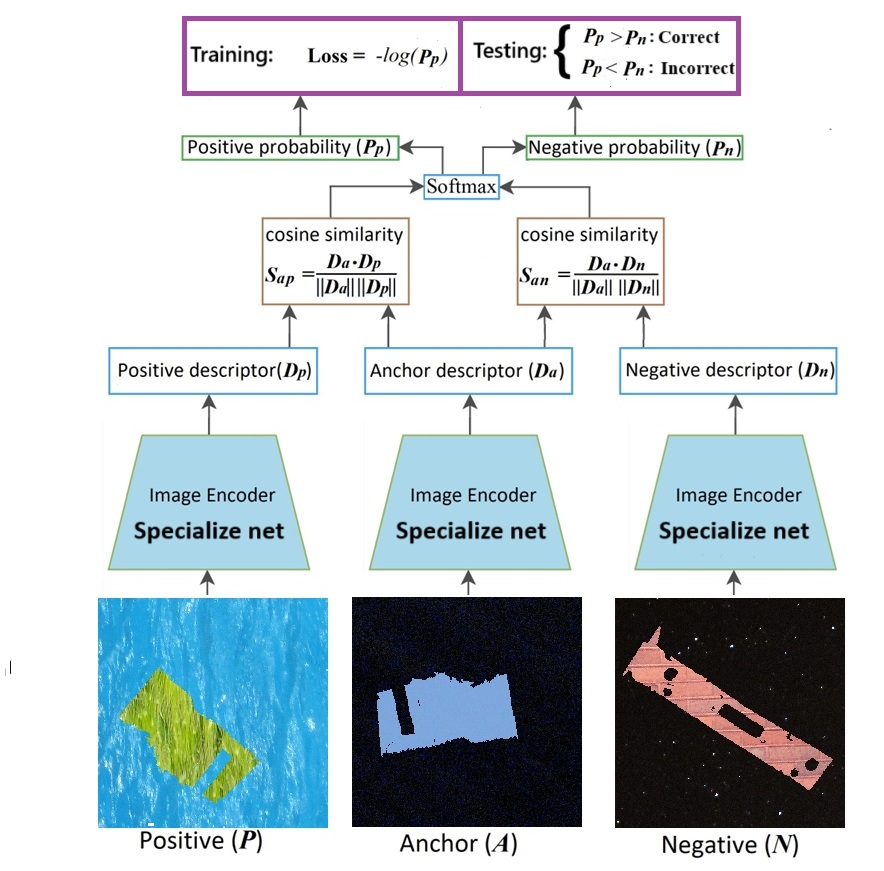}
  \vspace{-1.5em}
  \caption{\textbf{Training specialized net for shape texture
 recognition/retrieval:} Neural nets for matching
shapes, textures, and materials were trained using contrastive learning. 
\newline
a) Each training/testing batch involves choosing two images for each instance (shape or texture),
one of these images was considered as the anchor the other as positive. 
\newline
b) The embedding/descriptor vectors of these images were extracted using the net (Convnext Base). 
\newline
c) In addition, several images
of different instances (negatives) were chosen and their embedding was extracted using the same
net. 
\newline
d) The cosine similarity between all descriptors and the descriptor of the anchor was calculated
and passed through a softmax function to calculate the match probabilities between each pair of images. These
probabilities were used to calculate cross-entropy loss (with the anchor-positive probability as the
true class). 
\newline
The training was done (using a single RTX 3090) independently on each of the LAS\&T domains (2D shapes, 3D
shapes, 2D textures, 3D materials). Each set with all possible variations of the instance including size. For 3D cases an aggressive image augmentation where used including noise, gaussian
blurring, rotation, decoloring, histogram equilibration, and various camera effects. The aggressive
augmentation was necessary to cross the synthetic to real domain gap and allow the model to maintain
high accuracy on the real image benchmarks (without augmentation there were significant drops
when testing on real images.). For testing 1 positive image and 2 negative images were used to match
the LVLM tests (the above image only shows 1 negative for simplicity).}
  \label{fig:speciallizenet}
\end{figure*}

\begin{figure*}[t]
  \centering
   \includegraphics[width=1\textwidth]{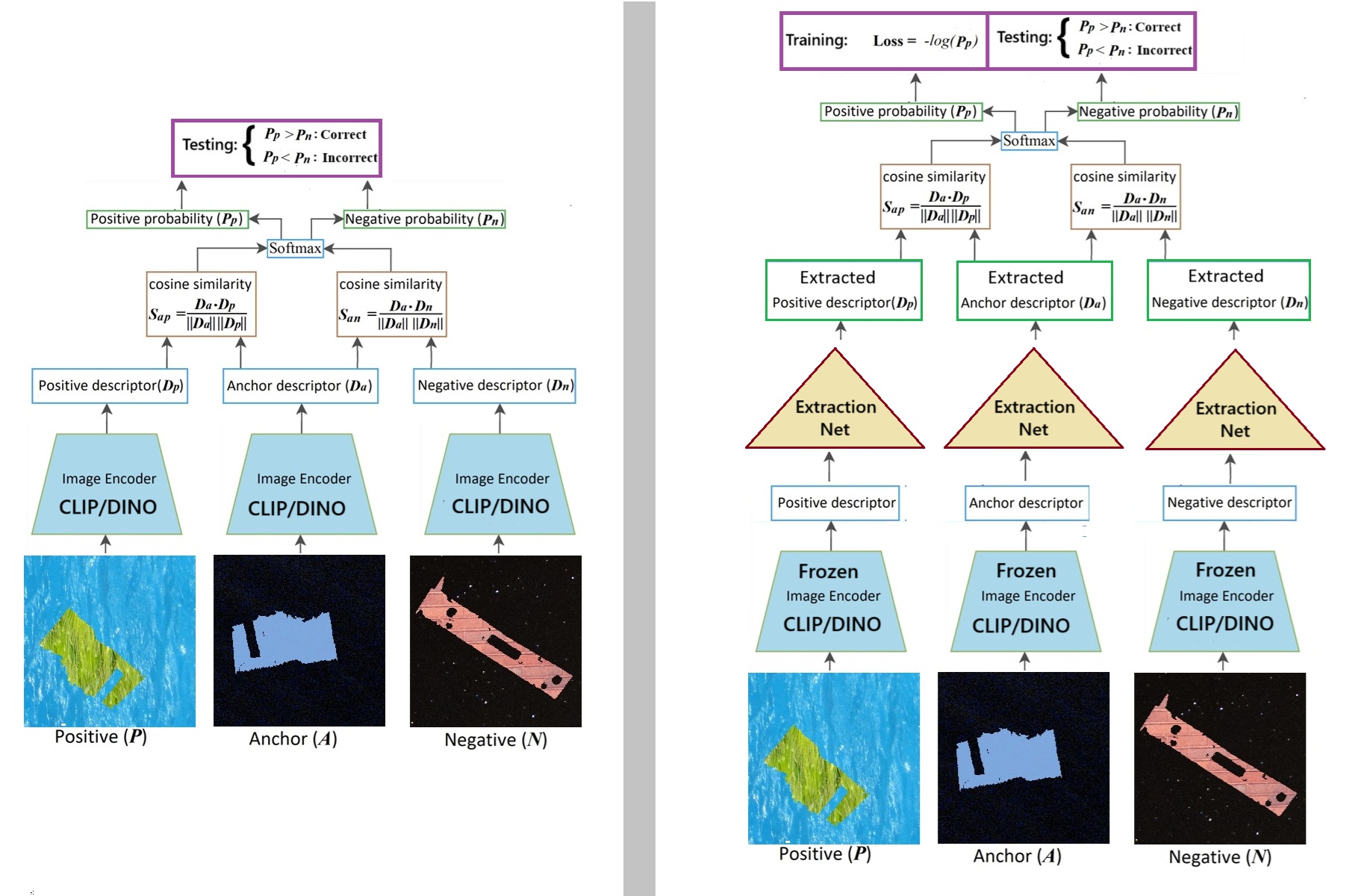}
  \vspace{-1.5em}
  \caption{\textbf{Left: Shape and texture matching using vision foundation model embeddings (CLIP/DINO)}. Using the embedding of pretrained open CLIP/Dino V2 to match shape and textures was done simply by taking the pretrained version of these models and running them on a two different images of the same instance (shape/texture)   and using pretrained DINO/CLIP to generate their embedding. One of these images was defined as an anchor and the other as positive. In addition, a few images of different random instances were also taken and processed to generate the embeddings, these were defined as negatives.   The cosine similarities between the embeddings of the anchor and the positive descriptor and the anchor and negative descriptors were calculated. If the similarity between the embeddings of images of the same instance (anchor and positive) was the highest of all pairs this was considered correct, else this was considered an error. Note that this is a shape-matching test with 1 positive (Correct) choice and 2 negative (incorrect choices) similar to the 4 panels test given to the LVLMs. No training was used.
  \newline
\textbf{Right: Refining CLIP/DINO descriptors:} To obtain task-specific descriptors from the pretrained CLIP or DINO models, we first extract the image embeddings using the frozen backbone. These embeddings are then passed through an additional, trainable multi-layer perceptron (MLP), which outputs refined descriptors tailored to shape or texture recognition. The MLP, typically composed of 4–10 layers (with minimal gains beyond 4 layers), is trained using a contrastive learning approach. Specifically, for each batch, the cosine similarity between the anchor and positive (same instance) descriptors serves as the logit for the positive class, while the similarities between the anchor and negative (different instance) descriptors serve as the logits for the negative classes. These logits are converted to matching probabilities via a Softmax function and used to compute the cross-entropy loss. Importantly, the CLIP/DINO backbone remains frozen throughout training, ensuring that only the MLP adapts to the shape or texture recognition task. Altogether 4 separate nets were trained (for 2d shapes, 3D shapes,  2d textures, and 3D materials). The nets were trained using sets of images with maximum variation between instances, meaning everything including size was changed between instances.}
  \label{fig:clipdino}
\end{figure*}

\section{LAS\&T dataset size and properties.}
The size and properties of the LAS\&T dataset are summarized in \cref{tab:dataset_stats}. Note that the reported numbers correspond to \textbf{unique/different} shapes, textures, categories, and environments. For 2D shapes, approximately 100,000 instances were used to generate the test images, while an additional 365,000 shapes were extracted and are provided alongside the dataset.   3D shapes from about 1150 categories were taken from the Objaverse dataset \cite{deitke2023objaverse}, 2D shapes, textures, and PBR materials are all extracted from natural image and are not clearly associated with a defined set of categories. Physically based rendering (PBR) materials and 2D textures were sourced from the VasTextures dataset \cite{eppel2024vastextures}, which spans a broad range of domains but does not include class labels. High-dynamic-range (HDRI) environments used for background illumination in 3D scenes were obtained from the Poly Haven repository \cite{polyhaven_hdri}.
 \begin{table*}[t]
\centering
\begin{tabular}{lccccc}
\toprule
\textbf{Category} & \textbf{Shapes} & \textbf{Textures} & \textbf{Images} & \textbf{Shape Classes} & \textbf{Environments/HDRI/Baclgrounds} \\
\midrule
3D Shapes & 25k & $\sim$100k & 100k & 1150 & 500 \\
2D Shapes & 100k/365k & $\sim$100k & 200k/365k &  & $\sim$100k \\
3D Materials & $\sim$100k & 30k & 100k & 1150 & 500 \\
2D Textures & $\sim$200k & 100k & 200k &  &  $\sim$100k\\
3D Shapes Real-World Images & 36 &  & 130 &  & $\sim$130  \\
\bottomrule
\end{tabular}
\caption{Size and properties of the LAS\&T dataset. Textures refer to PBR materials in 3D and 2D textures in 2D. Shapes are either 2D masks or 3D meshes. Categories are object categories (car, house) available only for 3D.}
\label{tab:dataset_stats}
\end{table*}

\section{Dataset Generation}

\subsection{Synthetic 3D scene generation}
\label{APP:3D scene generation for}
Synthetic 3D scenes for both material and shape recognition were automatically created using the Blender 4.2 Python interface with CYCLES rendering\cite{lupianiblender}. First, a random object was loaded from the Objaverse dataset\cite{deitke2023objaverse}. 
The materials and textures of the object were removed and replaced by a random physics-based rendering material (PBR). PBRs are textures that cover the object's surface and simulate the properties of real materials and the way they interact with light. The source of the PBR materials was the Vatextures dataset, which contains around 500,000 highly diverse PBR materials extracted from real-world images using unsupervised approaches\cite{eppel2024vastextures}. The objects were randomly rotated. A random HDRI image was loaded and used as both a 3D panoramic background and a 360-degree illumination source to the object.
The images were then rendered (using a single RTX 3090).  

\subsection{Synthetic 2D scene generation} 
\label{App:2D scene generation for}
2D scene is created by loading random shapes with random orientations. Random 2D texture images were loaded from the VasTextures dataset\cite{eppel2024vastextures} and used to cover the shape (\cref{fig:overview}), alternatively the shape was white or uniformly colored. The background remained black, uniformly colored, or was covered with another random texture. The shape orientation, position and size were also changed in some cases.

\clearpage 

\section{Models Responses: Sampled explanation given by models when asked to explain their answers. Each response refers to the image above it.}
The section contains various responses given by the models to various queries.
The image and the query are given as well. Note that the query appears once for all images of the same type of question. Meaning if the query doesnt appear next to the image it means its the same as in previous image.  For these responses, the models were specifically asked to explain their reasoning.

\section{2D Shapes: Samples of models responses and explanations}
\label{appendix:2Dshapes_model_response}
\textbf{Query: } Which of the panels contain an identical 2D shape  to the shape in panels A?\\
 Note that the shape might be covered with different texture and have a different orientation.\\
 Explain your answer in detail, including how you got to it and what distinguishing features you used.\\
 
\includegraphics[width=\linewidth]{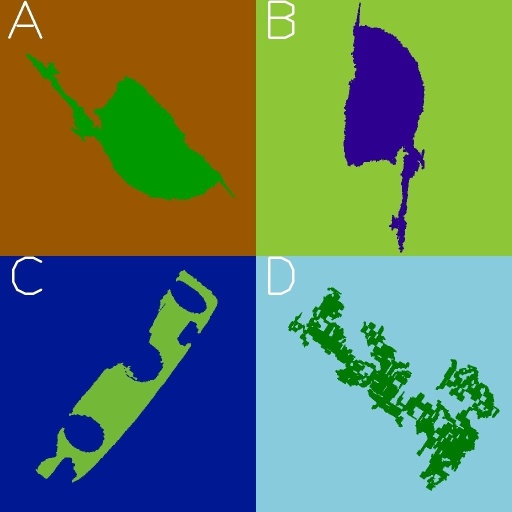} \\

\textbf{GPT 4O Response:  }\\
To determine which panel contains an identical 2D shape to the shape in panel A, we must focus on specific visual features and overall silhouette, disregarding orientation and texture.\\
1. **Panel A**: The shape is irregular, with a combination of curved and jagged edges. It has a distinct narrow protrusion extending from one end.\\
2. **Panel B**: The shape here doesn't match as it lacks the narrow protrusion and overall silhouette of panel A.\\
3. **Panel C**: The shape in this panel consists of a large hole and several projections, which do not match the shape's outline in panel A.\\
4. **Panel D**: This shape is fragmented and does not share the distinct narrow protrusion or overall silhouette seen in panel A.\\
\\
Hence, none of the shapes in panels B, C, or D match the shape in panel A. The answer is that none of the panels contain an identical 2D shape to the one in panel A.\\

\includegraphics[width=\linewidth]{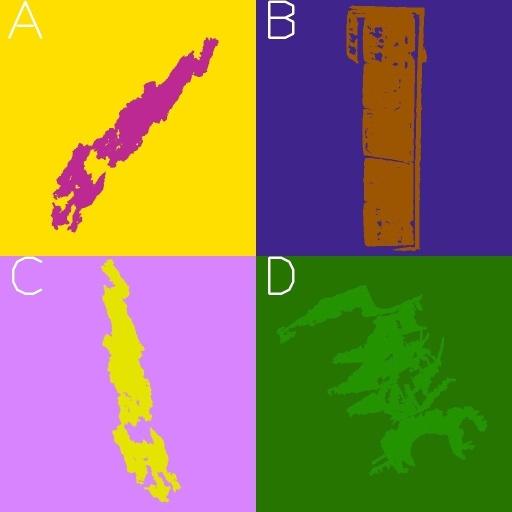} \\

\textbf{GPT 4O Response: }\\
The shape in panel A is identical to the shape in panel C.\\
\\
Here's how I identified it:\\
1. **Silhouette Matching**: The silhouette or outline of the shape in panel A and panel C is the same, even though the color and orientation differ. They both have a similar elongated, irregular structure with corresponding indents and protrusions.\\
2. **Texture and Orientation**: While the textures and colors are different, these don’t affect the actual shape. In panel A, the shape is magenta, and in C, it is yellow. The orientation in panel C is flipped compared to panel A, but the shape's fundamental structure remains identical.\\
3. **Distinctive Features**: Both shapes have a long, narrow body with a unique arrangement of protrusions that are distinct and easily matched when comparing the two panels.\\
\\
Panel B and D contain different shapes altogether, as their outlines do not match the shape in panel A.\\

\includegraphics[width=\linewidth]{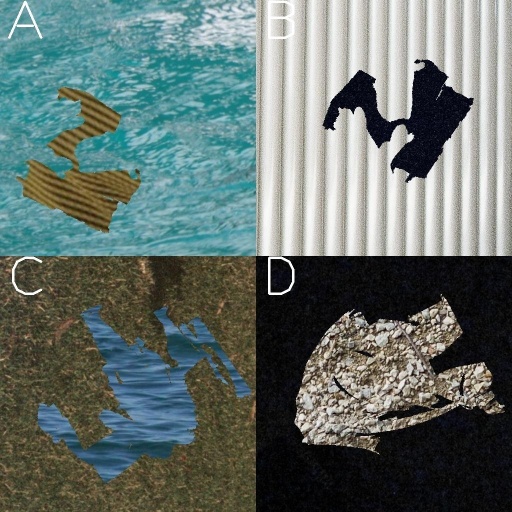} \\
\textbf{GPT 4O Response }\\
To identify the panel that contains an identical shape to the shape in panel A, we need to look at the outline and overall structure of the shape, regardless of the texture or orientation.\\
\\
1. **Panel A**: The shape has a very distinct silhouette with various curves and protrusions.\\
\\
2. **Panel B**: The shape here has different proportions and angles. It doesn’t match the shape in panel A.\\
\\
3. **Panel C**: The shape in this panel matches the shape in panel A. The distinctive features, such as the specific curves and angles, align perfectly when accounting for orientation differences. The texture and color are different, but the outline is identical.\\
\\
4. **Panel D**: The shape in panel D, although similar in some parts, does not completely match the distinct silhouette of the shape in panel A.\\
\\
Therefore, the shape in **panel C** is identical to the shape in panel A, considering the unique features of the outline and structure.\\

\includegraphics[width=\linewidth]{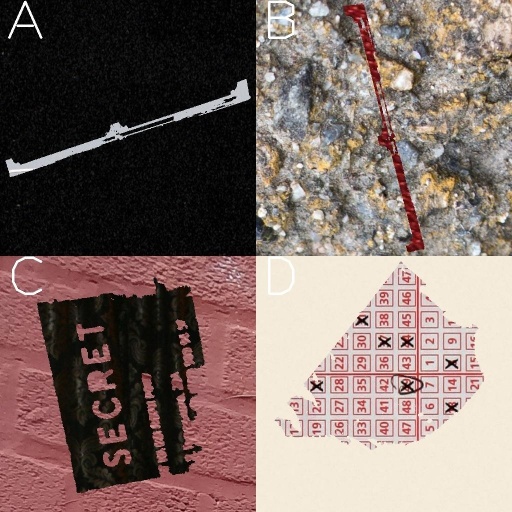} \\
\textbf{GPT 4O Response: }\\
The identical shape to the one in panel A is found in panel B.\\
\\
Here's the detailed explanation:\\
\\
- **Panel A**: The shape appears to be a long, thin form with a bar-like structure, having some intricate edges and segments sticking out.\\
 \\
- **Panel B**: This panel has a shape that matches Panel A quite closely in terms of overall outline and structure, despite being rotated and having a different texture. The long and thin nature with protrusions is consistent between panels A and B.\\
\\
- **Panel C**: This shape does not have the same long, thin structure and is more irregular and block-like, which does not resemble the shape in Panel A.\\
\\
- **Panel D**: This shape appears more triangular and irregular, lacking the linear and bar-like quality of Panel A.\\
\\
Due to the distinct elongated and segmented characteristics of the shape in Panel A, Panel B is the best match despite textural and orientation changes.\\
\\

\includegraphics[width=\linewidth]{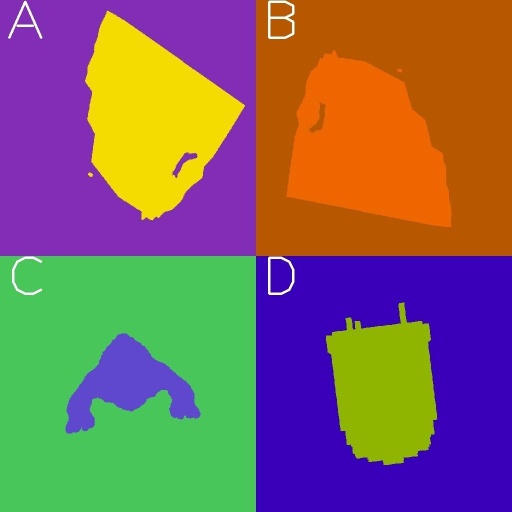} \\
\textbf{GPT 4O Response:  }\\
To identify which panel contains an identical 2D shape to the shape in panel A, we need to compare the overall outline and unique features of the shapes in each panel.\\
\\
1. **Panel A**: The shape has a distinctive outline with various protrusions and indentations. It appears asymmetrical and irregular, with a small cutout in the lower right section.\\
\\
2. **Panel B**: This shape has similar asymmetry and some indentations but lacks the distinctive cutout found in panel A. The overall structure appears different enough to rule this out.\\
\\
3. **Panel C**: The shape here is more symmetrical, resembling an arch or camel-like hump. It doesn't match the irregular, jagged edges of panel A.\\
\\
4. **Panel D**: This shape is closer in terms of the irregular and jagged outline, and it matches the orientation. Notably, this shape includes a small protrusion/cutout similar to what is seen in panel A.\\
\\
**Conclusion**: Panel D contains the shape identical to the one in panel A. The key distinguishing feature is the irregular outline and the small protrusion/cutout that matches between these two panels.

\includegraphics[width=\linewidth]{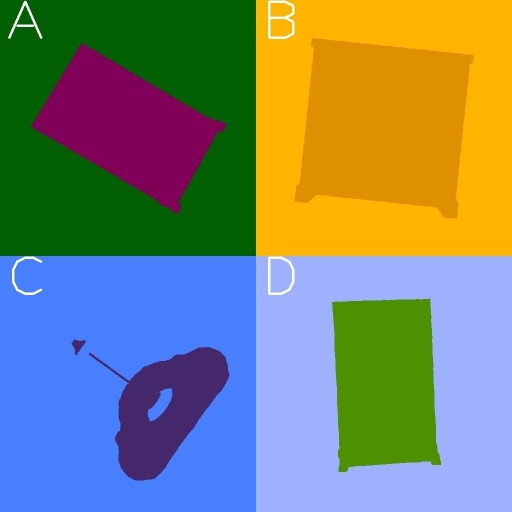} \\
\textbf{GPT 4O Response: }\\
\\
To find the panel with an identical shape to the one in panel A, we need to analyze each shape and compare their outlines.\\
\\
1. **Panel A**: Contains a rectangular shape with straight edges and four corners. It resembles a typical rectangle.\\
\\
2. **Panel B**: Also contains a rectangular shape, but with slightly detailed edges at two corners, making it more complex than the simple rectangle in panel A.\\
\\
3. **Panel C**: Contains an irregular shape that doesn't resemble a rectangle at all. It has curves and is not symmetrical or straight.\\
\\
4. **Panel D**: Contains a rectangular shape with straight edges and four corners, similar to the rectangle in panel A.\\
\\
**Conclusion**: The shape in panel D is identical to the shape in panel A. Both have a simple rectangular outline with straight edges and right angles, while panel B has additional detailing, and panel C is irregular.

\includegraphics[width=\linewidth]{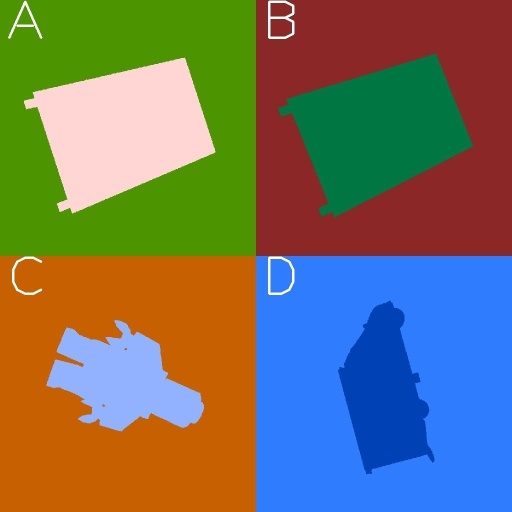} \\
\textbf{Gemini 1.5 pro: }\\
\textbf{Panel D contains the same shape as Panel A.}\\
\\
Here's how to determine this:\\
\\
1. **Focus on the overall outline:** The shape in Panel A is a quadrilateral (four-sided figure) that's somewhat like a parallelogram but with two short "tabs" or extensions on one of the longer sides.\\
\\
2. **Eliminate obvious mismatches:** Panels B and C have completely different shapes. Panel B is similar to A in that it is a quadrilateral with tabs, however, the angles of the quadrilateral are different from those in A.  Panel C has an altogether different shape.\\
\\
3. **Compare A and D:**  Though rotated and a different color, Panel D has the same quadrilateral shape as A, including the two small tabs on one of the long sides. If you were to rotate the shape in D counter-clockwise and expand it, it would overlay the shape in A.\\
\\
Therefore, only panel D truly duplicates the shape found in panel A.\\

\includegraphics[width=\linewidth]{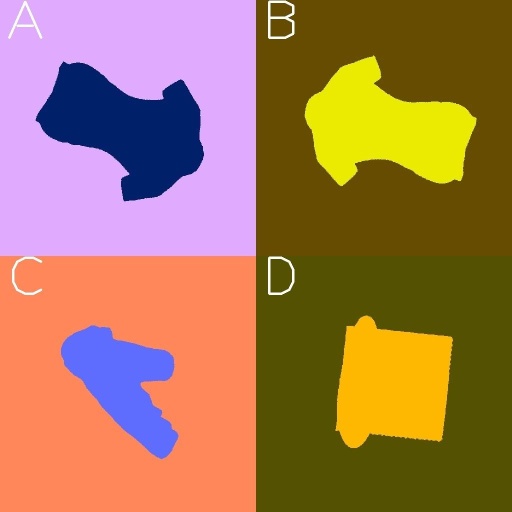} \\
\textbf{Gemini 1.5 pro:}\\
Panel B contains the same shape as panel A.\\
\\
Here's how to tell:\\
\\
1. **Visual Comparison and Mental Rotation:** While the colors and orientations are different, if you mentally rotate the shape in panel B roughly 90 degrees counterclockwise, you'll see it has the same overall outline and proportions as the shape in panel A.  They both appear to be simplified representations of a t-shirt.\\
\\
2. **Distinguishing Features:** Both shapes share key characteristics: a short, somewhat rounded body with a smaller, slightly curved protrusion on one side (representing a sleeve).  The lengths and relative positions of these features match up once you account for the rotation.  

\includegraphics[width=\linewidth]{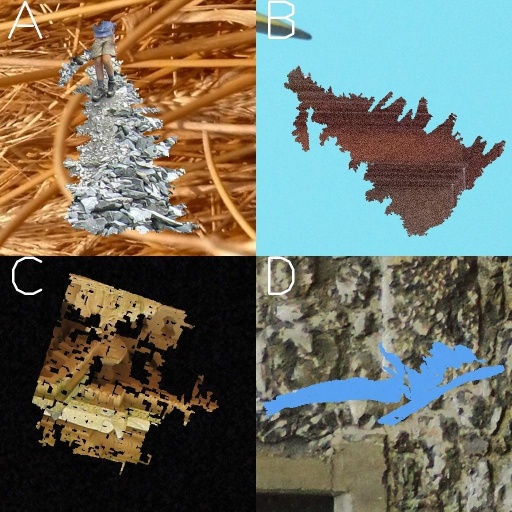} \\
\textbf{Gemini 1.5 pro: }\\
Panel C contains a rotated version of the 2D shape in Panel A.\\
\\
Here's why:\\
\\
1. **Outline Matching:** Both shapes share a distinctive, irregular outline.  While the textures are entirely different (rocks and trail in A,  parts of a building or statue in C), the outer boundary is essentially the same.\\
\\
2. **Rotation:** Panel C's shape is rotated clockwise compared to Panel A. If you were to mentally rotate C counter-clockwise, the alignment with A would become more evident.\\
\\
3. **Distinguishing Feature:**  Notice the prominent "arm" or branch-like extension sticking out towards the upper-right in A.  This same feature, albeit rotated, is present in C towards the bottom-right. This unique part of the silhouette confirms the shapes are identical, just differently oriented and textured.

\includegraphics[width=\linewidth]{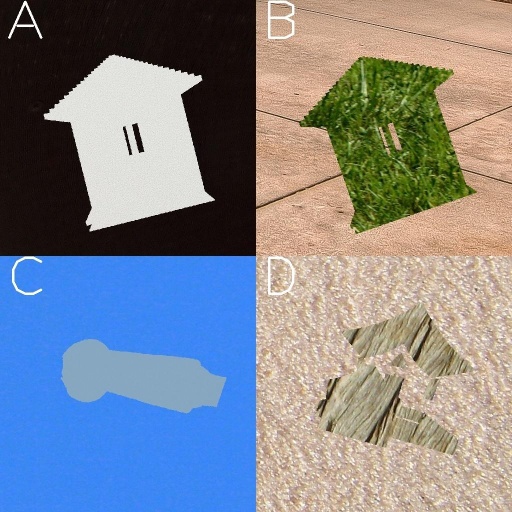} \\
\textbf{Gemini 1.5 pro:}\\
Panels **B, C, and D** all contain the same 2D shape as panel A.\\
\\
Here's the breakdown:\\
\\
1. **Panel A:** This shows a simple house-like shape with a pitched roof and two small rectangular windows.\\
\\
2. **Panel B:** Although covered with a grass texture and slightly rotated, the underlying shape is identical to panel A.  The roofline and the two window cutouts are clearly visible.\\
\\
3. **Panel C:** This panel has a slightly more abstract appearance.  The shape has been rotated and filled in with a solid light blue.  However, the outline still matches the house shape of panel A. Imagine tilting the house in panel A counter-clockwise, and you'll see the roofline and side wall match the silhouette in C.\\
\\
4. **Panel D:** This panel's shape is covered with a wood texture and appears to be slightly warped or distorted. However, the underlying outline still corresponds to the house shape. The roofline angles and the relative positions of the "window" areas (now filled with wood texture) match panel A.\\
\\
\\
The key is to look past the textures and rotations and focus on the fundamental 2D shape outlined by the edges of the figures.  All four panels maintain that same core structure.

\includegraphics[width=\linewidth]{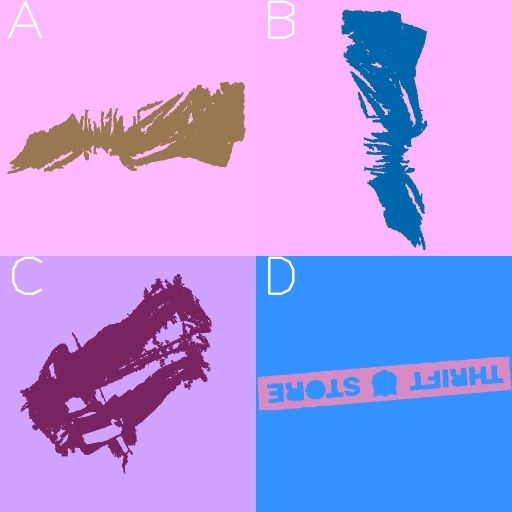} \\
\textbf{Gemini 1.5 pro: }\\
Panel C contains a shape identical to Panel A.\\
While the colors and textures are different, the overall outline and proportions of the shape are the same. It's as if the shape in Panel A has been rotated and mirrored. Imagine taking a tracing of the shape in Panel A, flipping it horizontally, and then rotating it counter-clockwise. It would align perfectly with the shape in Panel C.  The key features like the pointed end and the curved opposite side remain consistent.  B and D are different shapes altogether.
\includegraphics[width=\linewidth]{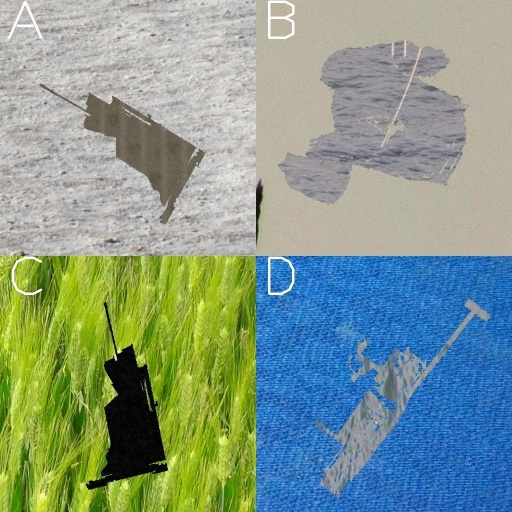} \\
\textbf{Grok 2 Vision: } To determine which panels contain an identical 2D shape to the shape in panel A, we need to compare the shapes while considering that they might be covered with different textures and have different orientations. Here’s the detailed analysis:\\
\\
1. **Panel A**: This is our reference shape. It has a distinct outline with several key features:\\
   - A long, thin protrusion extending from the left side.\\
   - A rectangular section with a cut-out on the right side.\\
   - A smaller protrusion on the top left.\\
\\
2. **Panel B**:\\
   - The shape in panel B has a different overall structure. It is more rounded and lacks the long, thin protrusion seen in panel A.\\
   - The texture is different, but the fundamental shape is not similar to panel A.\\
\\
3. **Panel C**:\\
   - The shape in panel C has a similar outline to panel A when considering rotation.\\
   - It has the long, thin protrusion on the left side (when rotated 90 degrees counterclockwise).\\
   - The rectangular section with the cut-out on the right side matches the shape in panel A.\\
   - The smaller protrusion on the top left is also present.\\
\\
4. **Panel D**:\\
   - The shape in panel D has a similar outline to panel A when considering rotation.\\
   - It has the long, thin protrusion on the right side (when rotated 90 degrees clockwise).\\
   - The rectangular section with the cut-out on the left side matches the shape in panel A.\\
   - The smaller protrusion on the top right is also present.\\
\\
**Conclusion**:\\
- Both panels C and D contain shapes that are identical to the shape in panel A, albeit rotated. Panel C is a 90-degree counterclockwise rotation of panel A, and panel D is a 90-degree clockwise rotation of panel A.\\
\\
**Distinguishing Features Used**:\\
- The long, thin protrusion.\\
- The rectangular section with a cut-out.\\
- The smaller protrusion on the opposite side of the long protrusion.\\
\\
Thus, the panels that contain an identical 2D shape to the shape in panel A are **panels C and D**.

\includegraphics[width=\linewidth]{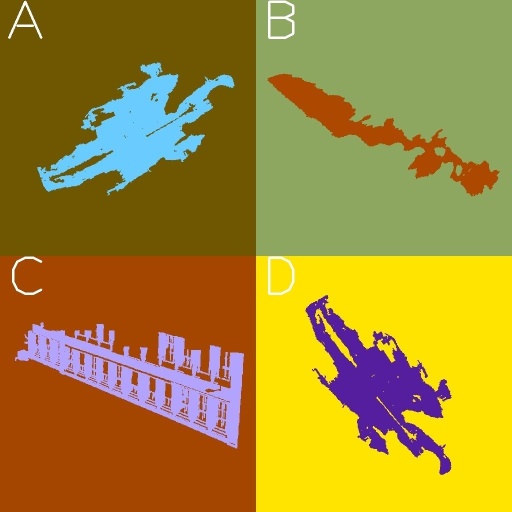} \\
\textbf{Grok 2 Vision }\\
To determine which panel contains an identical 2D shape to the shape in panel A, we need to compare the shapes based on their outlines, ignoring differences in texture, color, and orientation. Here’s a detailed analysis:\\
1. **Panel A**: This is our reference shape. It is a complex, irregular shape with multiple protrusions and indentations. Key features to note include:\\
   - The overall elongated shape with a central body and several branching extensions.\\
   - Specific notches and protrusions that create a unique silhouette.\\
2. **Panel B**:\\
   - This shape is also elongated and has a similar complexity with multiple branches and indentations.\\
   - Upon closer inspection, the outline of the shape in panel B matches the outline of the shape in panel A. The orientation is different, but the key features like the central body and the specific notches and protrusions are the same.\\
3. **Panel C**:\\
   - This shape appears to be a different type of structure, resembling a building facade with regular, repeating patterns of windows and columns.\\
   - The outline is fundamentally different from the shape in panel A, which is more organic and irregular.\\
4. **Panel D**:\\
   - This shape is similar in complexity to the shape in panel A, but upon closer inspection, the outline does not match. The branches and indentations are in different positions and the overall structure is different.\\
**Conclusion**: The shape in panel B is identical to the shape in panel A. The distinguishing features used in this comparison include:\\
- The overall elongated form with a central body.\\
- The specific arrangement of protrusions and indentations.\\
- Ignoring color, texture, and orientation to focus solely on the outline.\\
Therefore, **panel B** contains an identical 2D shape to the shape in panel A.

\includegraphics[width=\linewidth]{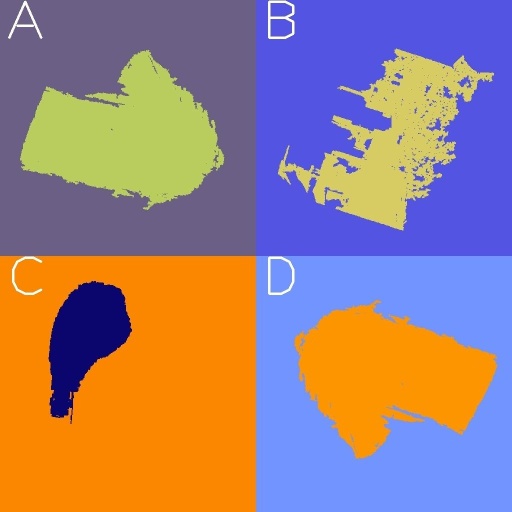} \\
\textbf{Grok 2 Vision \# Wrong use of disitingush features}\\
To determine which panels contain an identical 2D shape to the one in panel A, we need to compare the shapes based on their outlines, disregarding texture, color, and orientation. Here’s a detailed analysis:\\
1. **Panel A**: This is our reference shape. It has a distinct outline with several characteristic features:\\
   - A broad base on the left side.\\
   - A prominent protrusion on the top left.\\
   - A relatively smooth top edge.\\
   - A jagged right edge with multiple small inlets and outlets.\\
   - A small protrusion on the bottom right.\\
2. **Panel B**: This shape has a very different outline:\\
   - It is more fragmented with many internal inlets and islands.\\
   - The overall shape is more complex and less smooth compared to A.\\
   - The orientation and structure are distinctly different.\\
3. **Panel C**: This shape appears to match the outline of panel A when considering rotation:\\
   - The broad base on the left side is visible but rotated.\\
   - The prominent protrusion on the top left in A corresponds to the bottom left in C.\\
   - The smooth top edge in A matches the left edge in C.\\
   - The jagged right edge in A matches the top edge in C.\\
   - The small protrusion on the bottom right in A matches the top right in C.\\
4. **Panel D**: This shape does not match the outline of panel A:\\
   - It has a more elongated and less jagged outline.\\
   - The overall structure and proportions are different.\\
Based on this analysis, the shape in **Panel C** is identical to the shape in **Panel A** but rotated. The distinguishing features used to make this determination include the overall outline shape, the position of protrusions, and the jaggedness of certain edges.\\
\\
Thus, **Panel C** contains an identical 2D shape to the shape in Panel A.\\

\includegraphics[width=\linewidth]{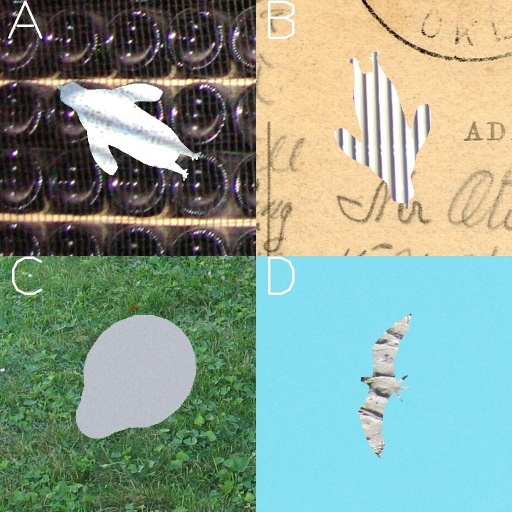} \\
\textbf{GPT 4O Response:}\\
The object in panel D has an identical 2D shape to the object in panel A.\\
\\
Here's how I determined that:\\
\\
1. **Shape Analysis**:\\
   - **Panel A**: The shape resembles a flying bird with wings extended, showing a distinct silhouette with two horizontal extensions (wings) and a pointed tail.\\
   - **Panel D**: The shape also resembles a flying bird, similar to the one in panel A, with the same wing extension and pointed tail features.\\
\\
2. **Orientation**:\\
   - Despite possible differences in orientation, the silhouettes in panels A and D align in terms of wing and tail structure.\\
\\
3. **Texture and Context**:\\
   - The texture and background are different in panel D compared to panel A, but the outline remains consistent.\\
\\
Panels B and C differ significantly in their outlines:\\
- **Panel B**: Has a more vertical, irregular, finger-like shape.\\
- **Panel C**: Exhibits a more rounded and blob-like shape without distinct wing or tail features.\\
\\
Thus, the matching object based purely on shape is in panel D.\\

\includegraphics[width=\linewidth]{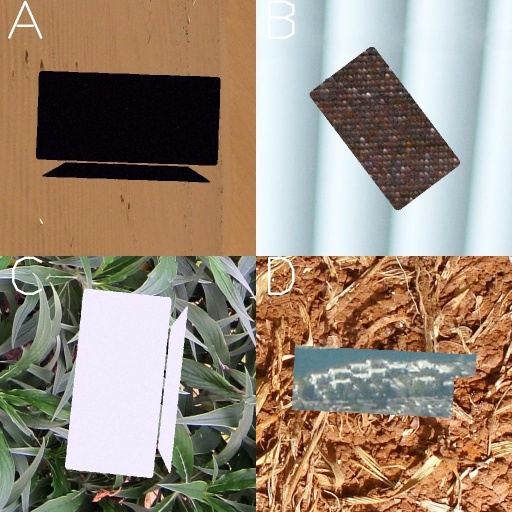} \\
\textbf{GPT 4O Response:}\\
To determine which panel contains an object with an identical 2D shape to the object in panel A, we need to analyze the silhouette in panel A and compare it to the silhouettes in panels B, C, and D.\\
\\
**Panel A:**\\
- Shape: The object has a shape resembling a trapezoid with a smaller trapezoid below it, creating a shape similar to a computer screen on a stand.\\
\\
**Panel B:**\\
- Shape: The object in panel B is a rectangle. This shape doesn't match the two-part trapezoidal shape in panel A.\\
\\
**Panel C:**\\
- Shape: The object in panel C is a rectangle, but it is shown in a three-dimensional orientation, appearing as a folded card. This doesn't match the silhouette shape in panel A.\\
\\
**Panel D:**\\
- Shape: The shape in panel D is a rectangle. It doesn't match the two-part trapezoid shape of the object from panel A.\\
\\
Based on the characteristics of the shape in panel A, none of the objects in panels B, C, or D appear to have an identical 2D shape. The primary feature is the combined trapezoid shape and none of the other panels have an equivalent shape with a stand.\\

\includegraphics[width=\linewidth]{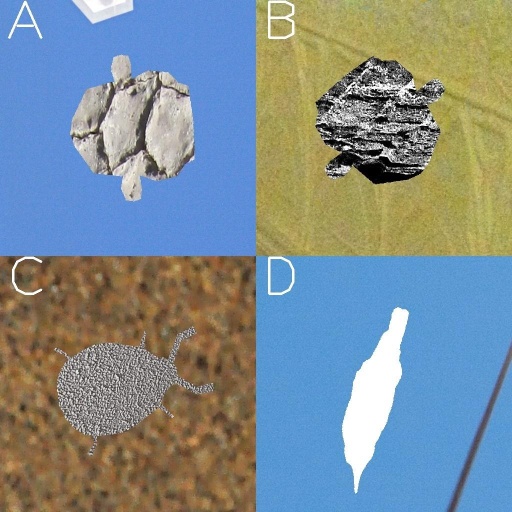} \\
\textbf{Gemini 1.5 pro respond:   }\\
Panel D contains an object with an identical 2D shape to the object in Panel A.\\
\\
Here's how to determine this:\\
\\
1. **Panel A:** Shows a turtle shape with a distinct pattern resembling stone or rock.  Focus on the overall outline of the turtle, including its head, legs, and shell.\\
\\
2. **Panel B:** While somewhat similar, this turtle shape is significantly different.  Notice the smoother contours and less-defined limbs.  The pattern is also distinct.\\
\\
3. **Panel C:**  This is an insect shape, clearly not a match to the turtle.\\
\\
4. **Panel D:** This shape, although rotated and plain white, matches the outline of the turtle in Panel A. If you mentally rotate the shape in Panel D clockwise by approximately 45 degrees, it aligns with the turtle's orientation in Panel A. The head, legs, and shell contours match. The only differences are the rotation, color, and lack of texture.\\
Therefore, only Panel D shares the identical 2D shape with Panel A.\\

\includegraphics[width=\linewidth]{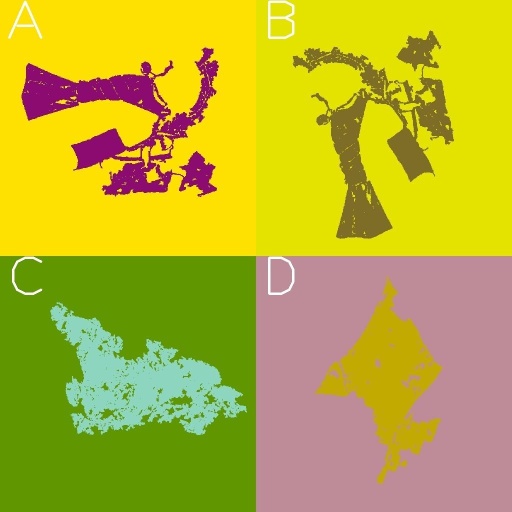} \\
\textbf{Grok 2 vision response:  }\\
To determine which panels contain an object with an identical 2D shape to the object in panel A, we need to analyze the shapes, ignoring texture, color, and orientation. Here’s a detailed breakdown of each panel:\\
1. **Panel A**:\\
   - Contains a complex shape with distinct features such as:\\
 	- A large, broad section on the left.\\
 	- A narrow, elongated section extending to the right with a few protrusions.\\
 	- A smaller, rectangular section below the main body.\\
 	- Several smaller, irregularly shaped sections connected to the main body.\\
2. **Panel B**:\\
   - The shape here is quite different from A. It has:\\
 	- A central, vertical elongated section.\\
 	- Several smaller, irregular sections branching off from the central part.\\
 	- A distinct, separate rectangular section on the right.\\
   - The overall structure and arrangement of parts do not match panel A.\\
3. **Panel C**:\\
   - This shape is also different from A. It has:\\
 	- A more compact, irregular outline.\\
 	- Lacks the distinct large, broad section on the left.\\
 	- Does not have the same elongated narrow section extending to the right.\\
   - The overall shape and arrangement of parts do not match panel A.\\
4. **Panel D**:\\
   - This shape is different from A as well. It has:\\
 	- A more triangular overall shape.\\
 	- Lacks the large, broad section on the left.\\
 	- Does not have the same elongated narrow section extending to the right.\\
   - The overall structure and arrangement of parts do not match panel A.\\
\\
**Conclusion**:\\
After comparing the shapes in panels B, C, and D with the shape in panel A, it is clear that none of them have an identical 2D shape to the object in panel A. The distinguishing features such as the large, broad section on the left, the elongated narrow section extending to the right, and the specific arrangement of smaller sections are unique to panel A and not replicated in any of the other panels.\\
Therefore, **none of the panels (B, C, or D) contain an object with an identical 2D shape to the object in panel A**.

\includegraphics[width=\linewidth]{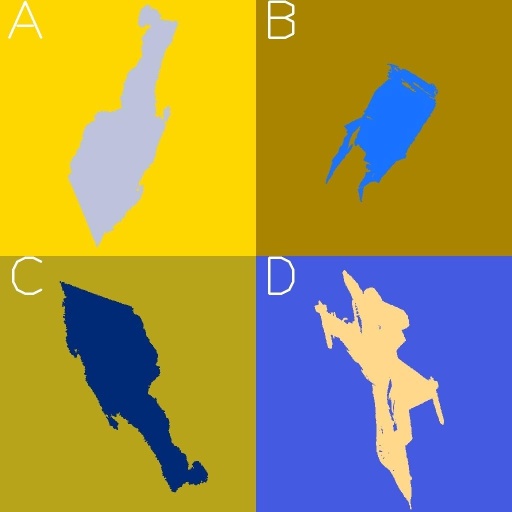} \\
\textbf{Grok 2 vision response:} \\
To determine which panels contain an object with an identical 2D shape to the object in panel A, we need to compare the shapes while ignoring differences in texture, color, and orientation. Here’s a detailed analysis of each panel:\\
\\
1. **Panel A**: This is our reference shape. It is a light blue shape on a yellow background.\\
2. **Panel B**: The shape here is blue on a brown background. To compare, we need to mentally rotate the shape in panel B to see if it matches the shape in panel A. Upon rotation, the shape in panel B appears to be identical to the shape in panel A, just rotated.\\
3. **Panel C**: The shape here is dark blue on a brown background. Comparing the outline, it is clear that the shape in panel C is significantly different from the shape in panel A. It has a more elongated and curved form.\\
\\
4. **Panel D**: The shape here is light yellow on a blue background. Again, we need to consider rotation. Upon careful inspection and mental rotation, the shape in panel D does not match the shape in panel A. The outline and proportions are different.\\
\\
**Conclusion**: The only panel that contains an object with an identical 2D shape to the object in panel A is **Panel B**. The distinguishing features used were the overall outline and proportions of the shapes, ensuring that rotation was considered in the comparison.\\

\section{3D Shapes: Samples of models responses and explanations. }
\label{appendix:3Dshapes_model_response}
\textbf{Query: “Which panel contains an object that has identical 3D shape to the object in panel A, but different in orientation and texture. Explain”.}

\includegraphics[width=\linewidth]{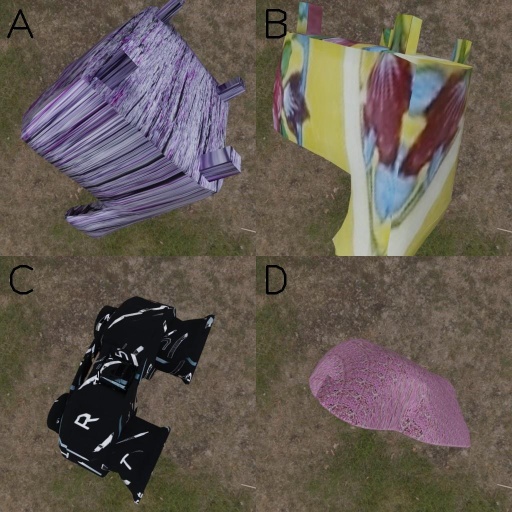} \\
   \textbf{GPT 4o response:} “Image D is the same shape as A, just rotated, smaller, and with a different texture. It's like taking the object from A, covering it with a pink, bubbly material, and then turning it over.”\\

\includegraphics[width=\linewidth]{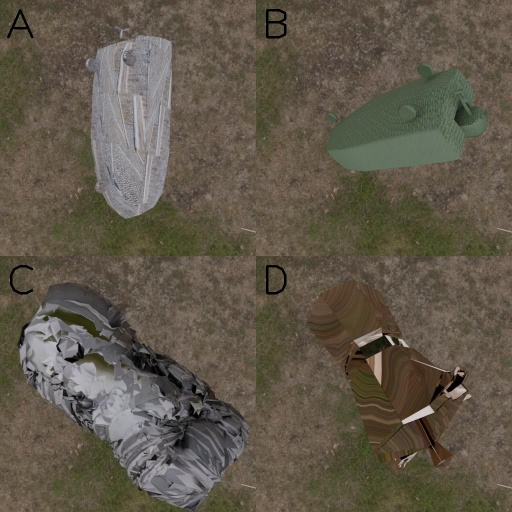} \\
\textbf{GPT 4o response:} Panel **D** is the same shape as A, just rotated and with a different texture. It appears to be made of polished wood, while A seems to be made of some sort of woven or layered material.”\\

\includegraphics[width=\linewidth]{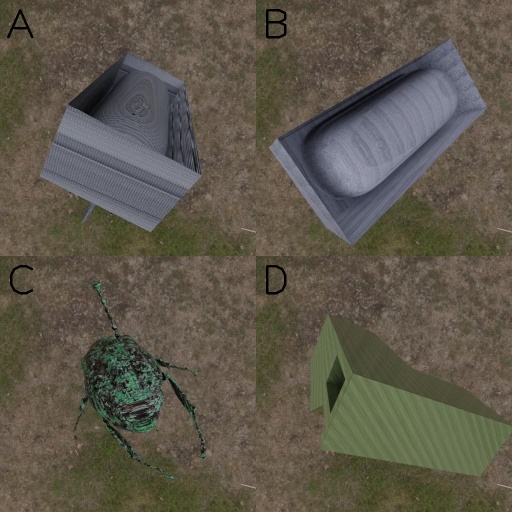} \\
\textbf{GPT 4o response:} “Panel D is the match for A.  It's the same basic triangular prism shape, just rotated and with a different (wood-like) texture applied.  Panel B is a rectangular prism with a curved indentation, and C is an insect”\\

\includegraphics[width=\linewidth]{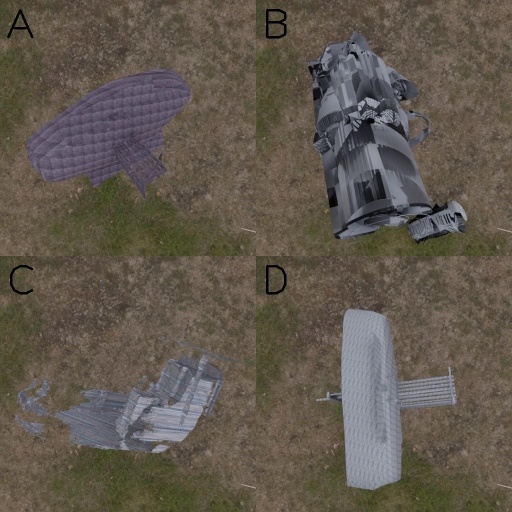} \\
\textbf{GPT 4o response (correct):} “Panel **D** shows the object that matches panel A's shape.  It's clearly the same basic form (a rounded, dish-like object with an attached structure), but it's rotated and has a completely different texture/material appearance. It also seems to be mirrored from left to right.” 
\includegraphics[width=\linewidth]{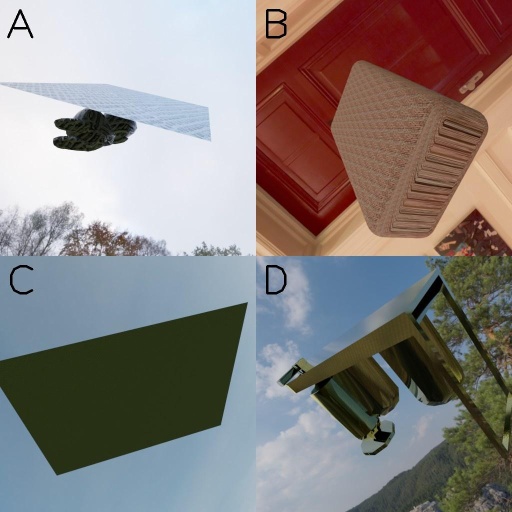} \\
\textbf{Querry: “Which panel in the image contains an object similar in 3D shape to the object in panel A, explain”. }\newline \textbf{Gemini response (Correct):} “The object in panel C is similar in 3D shape to the object in panel A. Both objects resemble a flat or planar structure. While panel A shows the object from a slightly angled underside view, panel C shows a flat, dark shape without detailing, indicating a similar planar form.”\\

\includegraphics[width=\linewidth]{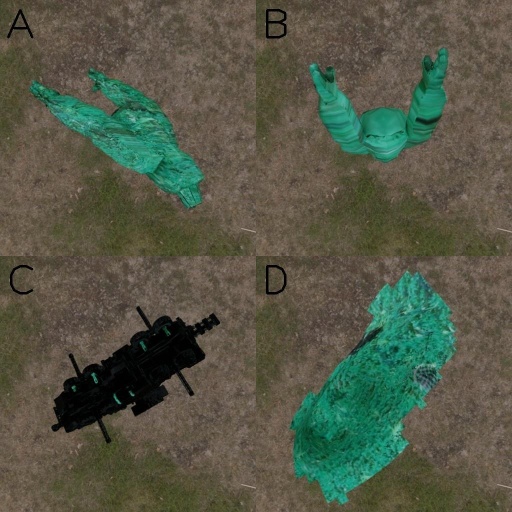} \\
\textbf{Gemini response:} “ Certainly! Let's analyze the shapes in the image.\\
**Analysis:**\\
* **Panel A:** Shows a green, somewhat elongated, and distorted humanoid-like figure. It appears to be lying on the ground.* **Panel B:** Shows a green humanoid-like figure standing straight with its arms raised. It is similar to A but not identical, as it's upright and the pose is different.\\
* **Panel C:** Shows a black metallic object, likely a machine or weapon, which bears no resemblance to the other objects.\\
* **Panel D:** Shows a green shape that appears to be a distorted version of the same humanoid-like figure as in panel A. It is leaning slightly.\\
**Conclusion:**\\
The panel that contains an object with an identical 3D shape to the object in panel A is **panel D**.\\
**Reasoning:**\\
Panel D's object clearly has the same structure and texture as panel A, it is only seen from a different angle and it appears to have rotated and tilted compared to panel A. Panel B and C are distinctly different objects and shapes.”\\

\includegraphics[width=\linewidth]{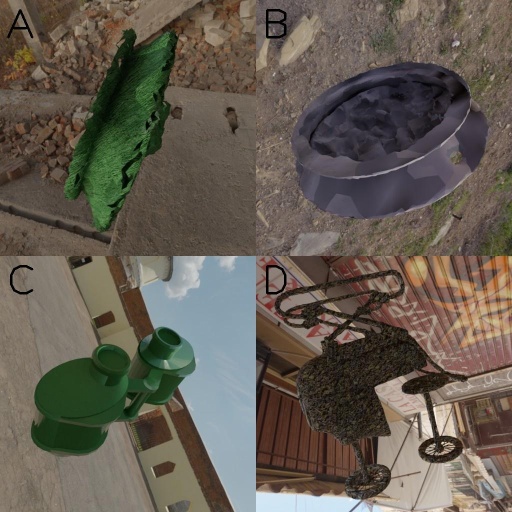} \\
\textbf{Gemini's response:}\\ “The object in panel A appears to be an irregular, elongated shape. The panel containing a similar 3D shape is likely panel D. Both objects have elongated, complex structures, though they are not identical. \\

\includegraphics[width=\linewidth]{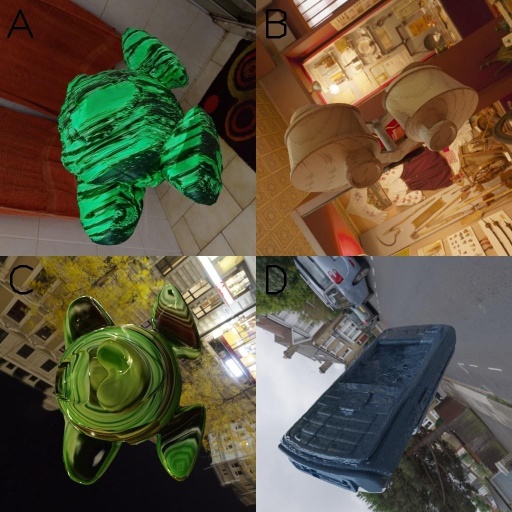} \\
\textbf{Query: “Which of the panels contain an object with identical 3d shape  to the object in panel A? Explain your answer.”}\\
Claude Response: “ Looking at all panels:\\
Panel A shows a shiny green balloon-like or inflated object with a lumpy, irregular shape.\\
Panel B shows mushroom-like shapes in beige/white colors.\\
Panel C shows a shiny green figure that appears to be a cartoon character or mascot sculpture with rounded shapes and ears.\\
Panel D shows what appears to be a dark-colored vehicle or bus from an unusual angle.\\
None of the other panels (B, C, or D) contain objects with identical 3D shapes to the irregular, lumpy form shown in Panel A. While Panel C is also green and shiny, it has a very different, more structured and character-like form compared to the abstract, bulbous shape in Panel A. Panels B and D show completely different types of objects with very different geometries.\\
Therefore, none of the other panels contain objects with identical 3D shapes to Panel A.”\\

\includegraphics[width=\linewidth]{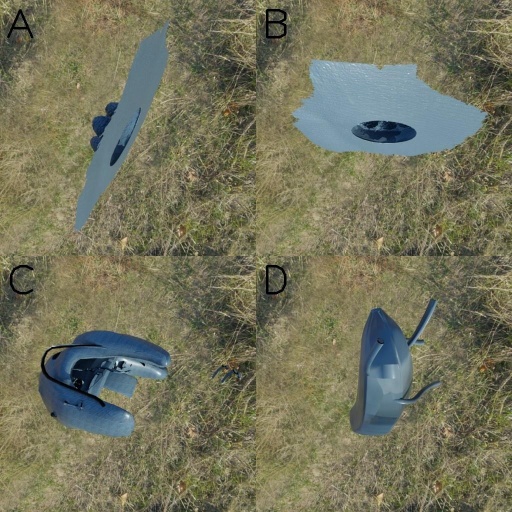} \\
\textbf{Query: “Which of the panels contain an object with identical 3d shape  to the object in panel A? Explain your answer.”}\\
Claude Response: “ Let me analyze each panel:\\
Panel A shows a narrow, elongated object with a gray/blue coloring.\\
Looking at the other panels:\\
- Panel B shows a wider, more rounded shape that appears quite different from A\\
- Panel C shows a curved, U-shaped object that has a very different form than A\\
- Panel D shows an object that appears to be the same as A but from a different angle, maintaining the same elongated form and proportions\\
Therefore, Panel D contains the object with identical 3D shape to Panel A. They appear to be the same object photographed from different viewpoints - Panel A shows a side view while Panel D shows a perspective view, but the fundamental 3D geometry remains the same. Panels B and C show distinctly different shapes that could not be the same object viewed from different angles.\\
The answer is Panel D.”  

\includegraphics[width=\linewidth]{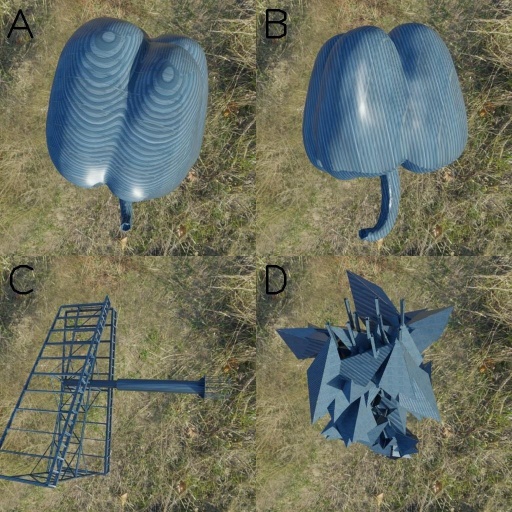} \textbf{Query “Which of the panels contain an object with identical 3d shape  to the object in panel A? Explain your answer.” } LLAma response: “\\
**Identical 3D Shapes in Panels B and C**\\
\\
Upon examining the image, it is evident that panels B and C contain objects with identical 3D shapes as the object in panel A.\\
\\
**Reasoning:**\\
*   The object in panel A has a distinctive shape with a curved top and a narrow base.\\
*   Panel B displays a similar object with the same curved top and narrow base, indicating that they share the same 3D shape.\\
*   In panel C, the object is oriented differently, but its curved top and narrow base are still visible, confirming that it also shares the same 3D shape as the object in panel A.\\
\\
**Conclusion:**\\
\\
Based on the visual evidence, panels B and C contain objects with identical 3D shapes as the object in panel A.”\\

\includegraphics[width=\linewidth]{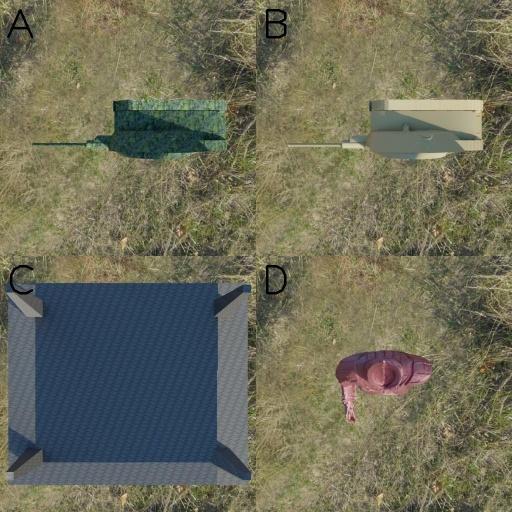} \\
\textbf{Query “Which of the panels contain an object with identical 3d shape  to the object in panel A? Explain your answer.” } LLAma response: “\\
The object in panel A is a tank. The object in panel D is also a tank, but it appears to be a different model or type of tank. The object in panel B is not a tank, but rather a piece of artillery or a cannon. The object in panel C is not a tank, but rather a building or a structure.\\
Therefore, the panels that contain an object with an identical 3D shape to the object in panel A are panels A and D.\\

\includegraphics[width=\linewidth]{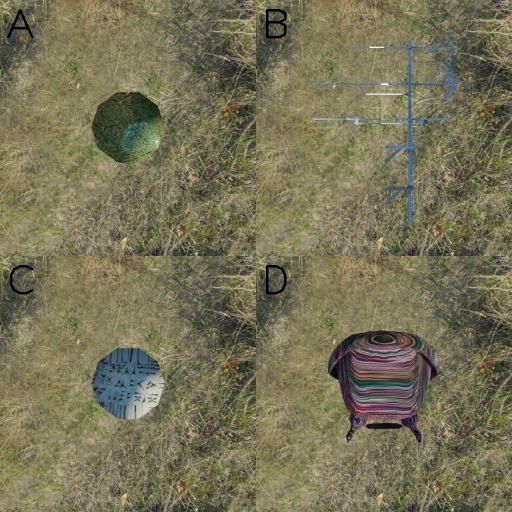} \\
LLama response: Panel A shows a three-dimensional object that resembles an octahedron. The same object appears in Panels C and D. However, Panel B shows an abstract blue figure with arrows and white bars, so it is not real. Answer: Panels A and D.

\includegraphics[width=\linewidth]{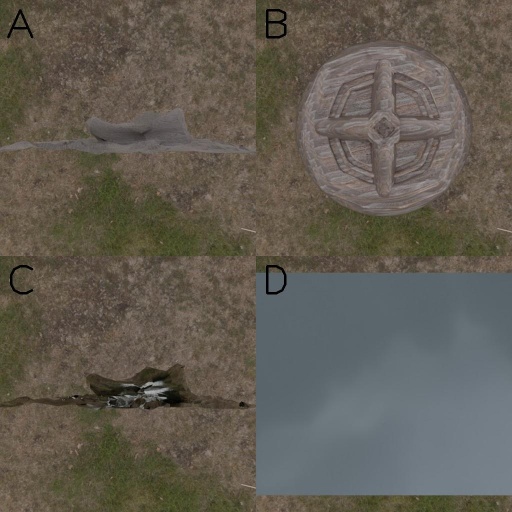} \\
  \textbf{Query:  “Which of the panels contains an object with an identical 3d shape to the object in panel A but with a different texture/material?”}  Claude's response:  “B - The shape in panel B appears to be the same draped/folded form as in panel A, but with a different texture that has a carved/embossed cross or compass-like pattern on it, while panel A shows a plain gray material.”\\

\includegraphics[width=\linewidth]{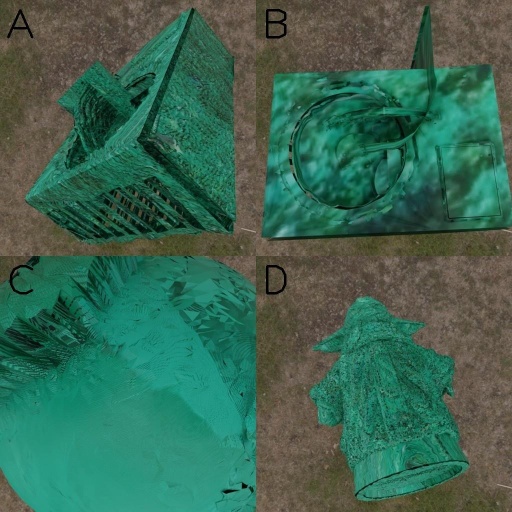} \\
\textbf{Gemini Response:} “ Based on the images provided, none of the panels (B, C, or D) contain an object with the exact same 3D shape as the object in panel A.\\
**Explanation:**\\
* **Panel A:**  Depicts what appears to be a rectangular building structure with a slanted roof section. There are also visible windows in the structure. The surface looks rough and textured.\\
* **Panel B:** Shows an object with a complex circular structure and a vertical element, resembling a sundial or similar device. It's quite different from the rectangular shape in A.\\
* **Panel C:** Presents a round, dome-like object. The surface is textured, but there are no straight edges, unlike panel A.\\
* **Panel D:** Depicts what appears to be a standing figurine with a cloak and pointed ears, likely based on the character Yoda. This is completely different in form to the building of panel A. **In summary:** The key to determining that there aren't identical 3D shapes is that while they share similar textures or colors (green), the basic forms of the objects in panels B, C, and D are significantly different from the rectangular building with a slanted roof as seen in panel A.”

\includegraphics[width=\linewidth]{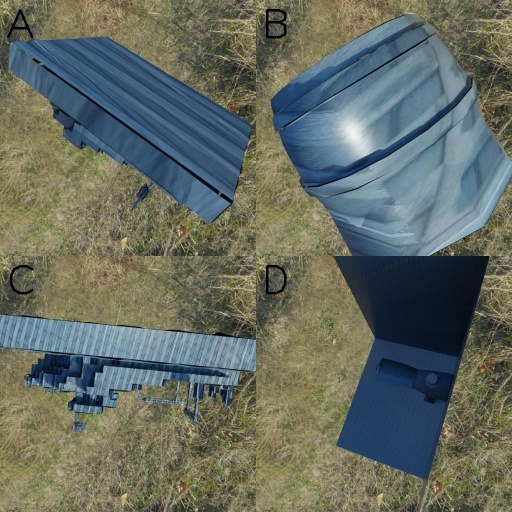} \\
\textbf{GPT 4O Response:} “The object in panel D has an identical 3D shape to the object in panel A. Both objects have similar flat, elongated rectangular forms without any visible extrusions or uneven surfaces. The other panels contain objects with distinct shapes, such as the cylindrical form in panel B and the jagged, irregular form in panel C.” \\

\newpage
\section{2D Textures: Samples of models responses and explanations}
\label{appendix:2D_texture model Respons}
\textbf{Query: “Which of the panels contain an identical 2D material/texture  to the shape in panels A?}\\
\textbf{ Note that the shape the texture appears on and the background might be different. The texture might also be rotated.  Explain your answer in detail, including how you got to it and what distinguishing features you used.”}

\includegraphics[width=\linewidth]{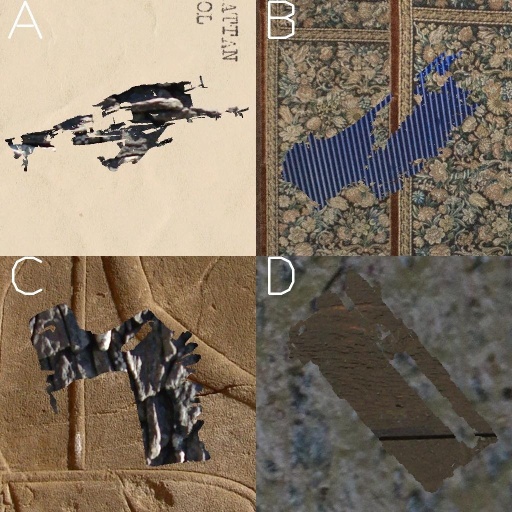} \\
\textbf{GPT 4o Response:} \\
Panel C contains an identical 2D material/texture to the shape in panel A.\\
\\
To determine this, I looked at several distinguishing features:\\
\\
1. **Color and Reflectiveness**: The texture in panels A and C has a metallic, reflective quality with a pattern of shades varying from dark to light.\\
\\
2. **Surface Pattern**: The texture has an uneven, crumpled appearance with complex shapes and creases, which matches between panels A and C.\\
\\
3. **General Shape Arrangement**: Although the shape and background can differ, recognizing the texture requires focusing on the surface pattern and reflectiveness rather than shape alone.\\
\\
Panel B shows a fabric-like pattern which is distinctly different, and panel D has a more uniform, less reflective texture. Therefore, panel C is the match.\\

\includegraphics[width=\linewidth]{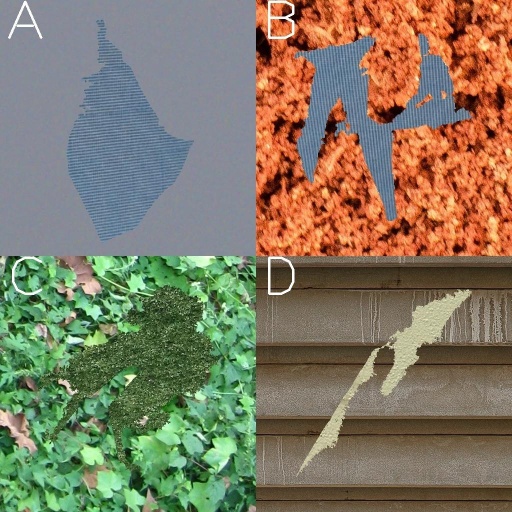} \\
\textbf{GPT 4o Response: }\\
To determine which panel contains an identical 2D material/texture to the shape in panel A, we need to examine the texture pattern closely and compare it to the other panels.\\
\\
**Panel A:**\\
- The texture is a uniform pattern of horizontal, closely spaced lines.\\
\\
**Comparison:**\\
\\
**Panel B:**\\
- The shape displays a similar blue pattern with horizontal lines. Although the background and shape are different, the texture matches perfectly.\\
\\
**Panel C:**\\
- The texture here is made up of a dense cluster of what appears to be grass or leaves, which is very different from the linear pattern in A.\\
\\
**Panel D:**\\
- The texture in this panel is a rough, stucco-like material, bearing no resemblance to the neat, linear pattern in A.\\
\\
**Conclusion:**\\
- Panel B contains the identical 2D texture to panel A. The distinguishing feature used to identify this match was the consistent horizontal line pattern that appears in both panels A and B, despite any rotation or difference in shape.\\

\includegraphics[width=\linewidth]{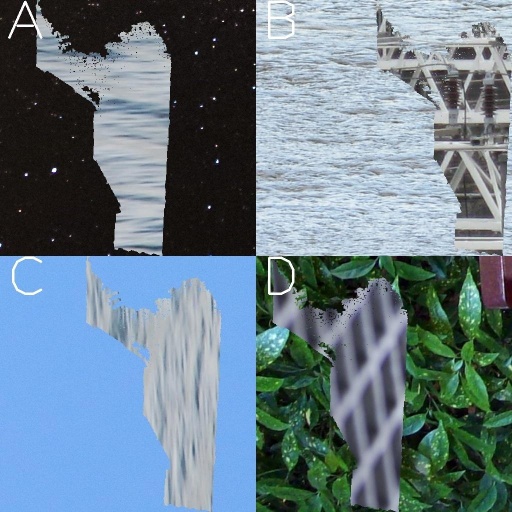} \\
\textbf{GPT 4o Response: } \\
To determine which panel contains an identical 2D texture to the shape in Panel A, we need to closely examine the texture features such as color, pattern, and texture details.\\
\\
Panel A depicts a water-like texture with horizontal dark and light stripes, resembling ripples on water.\\
\\
- **Panel B:** Contains a similar water-like texture, but the pattern seems to be different, with more pronounced and uniform lines.\\
- **Panel C:** Shows a vertical blue and white streak pattern, distinct from A’s horizontal pattern.\\
- **Panel D:** Displays a blurred, mesh-like pattern with a different color scheme and texture.\\
\\
The texture in **Panel A** and **Panel B** are both water-like, but with differences in the pattern density and orientation. The texture in **Panel C** is not only oriented vertically but also appears too blurred at the top to be identical to A.\\
\\
The water ripple texture in **Panel B** is horizontally aligned and more similar to A's pattern, but upon close inspection, the specific pattern of light and dark lines doesn't match exactly.\\
\\
Thus, although **Panel B** appears similar due to theme and design (water ripples), none of the panels contain a texture identical to **Panel A**. However, **Panel B** is the closest match in theme, despite variations in how the pattern is displayed.

\includegraphics[width=\linewidth]{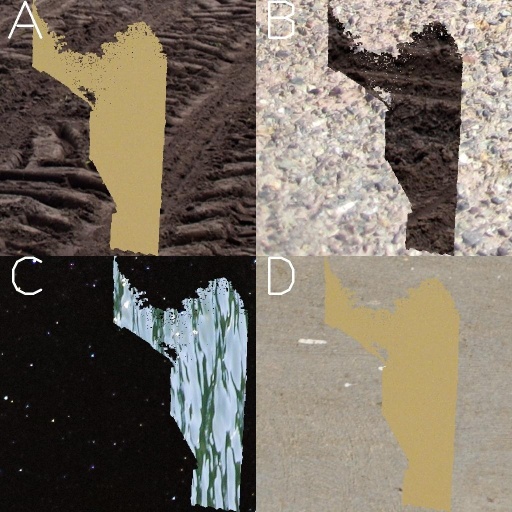} \\
\textbf{GPT 4o Response:  } \\
\#\#\# Analysis of Panel A:\\
- The texture features a series of parallel lines or ridges, suggesting a rough, possibly tread-like pattern.\\
- The color is a brown, earthy tone, indicative of soil or dirt.\\
\\
\#\#\# Examination of Other Panels:\\
\\
- **Panel B:**\\
  - The texture here looks like assembled bits or smaller pieces, not linear or ridged.\\
  - Color is similar to soil but lacks the distinct ridge pattern.\\
\\
- **Panel C:**\\
  - The texture appears as a flowing, watery or wavy pattern, not solid or ridged like treads.\\
  - Colors include various shades, none matching the earth tone of Panel A.\\
\\
- **Panel D:**\\
  - This texture features the same ridged pattern as Panel A.\\
  - The color also matches the brown, earthy hue.\\
\\
\#\#\# Conclusion:\\
Panel D contains the identical 2D texture to the shape in Panel A. The distinctive feature is the parallel, tread-like ridges with the same brown, soil-like color.\\

\includegraphics[width=\linewidth]{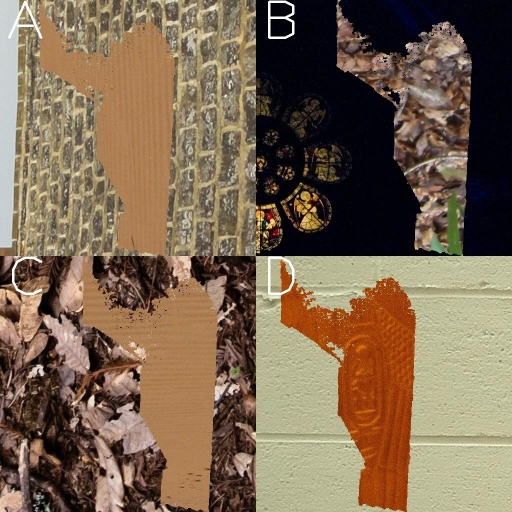} \\
\textbf{GPT 4o Response: } \\
The texture in Panel A is a cardboard-like texture with a linear, corrugated pattern. Let's analyze the other panels:\\
\\
- **Panel B**: This shows a multicolored, complex texture with floral or stained glass features. It does not match the simple linear pattern of Panel A.\\
\\
- **Panel C**: This panel has a similar cardboard-like texture with the same linear, corrugated pattern. The texture might be rotated but matches the one in Panel A.\\
\\
- **Panel D**: This includes a texture with a different pattern that looks like it has some symbols or designs, not the straightforward corrugation seen in Panel A.\\
\\
Therefore, **Panel C** contains an identical 2D texture to the shape in Panel A, distinguished by the same linear and corrugated pattern.\\

\includegraphics[width=\linewidth]{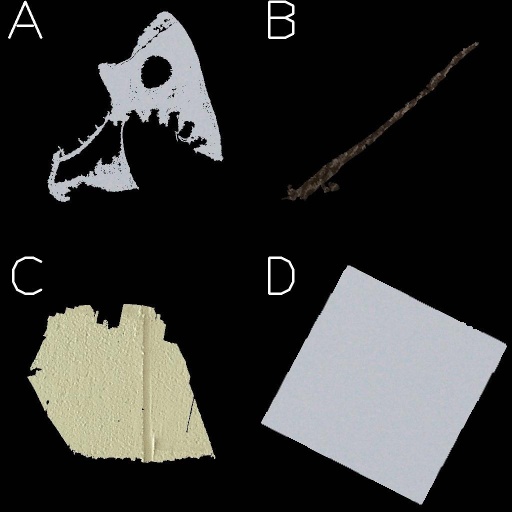} \\
\textbf{GPT 4o Response:}\\
To determine which panel contains an identical 2D texture to the shape in panel A, we need to analyze the specific texture features. The focus will be on the texture's color, pattern, and any distinctive marks or characteristics.\\
\\
\#\#\# Panel A\\
- **Texture Characteristics:** The texture in panel A is a grayscale texture with a certain roughness and jagged edges, resembling a fossil or bone with a speckled pattern.\\
\\
\#\#\# Panel B\\
- **Texture Characteristics:** Panel B features a long, thin, irregular texture that resembles a piece of driftwood or a thin stick. It does not match the speckled, bone-like texture of panel A.\\
\\
\#\#\# Panel C\\
- **Texture Characteristics:** The texture in panel C has a yellowish color and a surface that looks like a rough, stucco wall. The texture pattern and color are different from that of panel A.\\
\\
\#\#\# Panel D\\
- **Texture Characteristics:** The texture in panel D is a light-colored, smooth surface on a slanted square plane, resembling a piece of paper or fabric. The uniformity and smoothness differ significantly from the texture in panel A.\\
\\
\#\#\# Conclusion\\
None of the panels (B, C, or D) contain a texture identical to that of panel A. The texture in panel A has unique fossil-like characteristics that are not present in the other panels, which show distinct textures and colors.\\

\includegraphics[width=\linewidth]{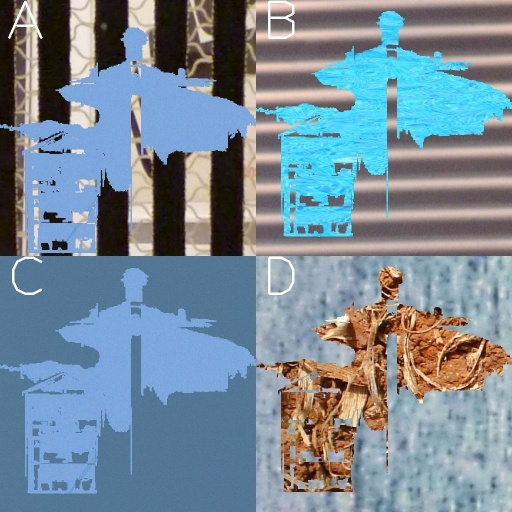} \\
\textbf{Grok 2 response:}\\
To determine which panels contain an identical 2D texture to the shape in panel A, we need to focus on the texture pattern within the shape, regardless of the shape's outline or the background. Here’s a detailed analysis:\\
1. **Panel A**: The texture within the shape appears to be a striped pattern with varying shades of blue, overlaid on a background with a more complex pattern.\\
\\
2. **Panel B**: The texture within the shape is also striped, similar to panel A. The stripes are horizontal and have a similar color gradient. The background is different, but the texture within the shape matches the pattern in panel A.\\
\\
3. **Panel C**: The texture within the shape is a solid blue color without any striped pattern. This is different from the texture in panel A.\\
\\
4. **Panel D**: The texture within the shape is a complex, organic pattern that resembles dried earth or roots, which is entirely different from the striped texture in panel A.\\
\\
\#\#\# Distinguishing Features:\\
- **Striped Pattern**: The key feature to look for is the striped pattern with varying shades of blue. Panels A and B both have this pattern.\\
- **Color Gradient**: The gradient of the stripes should be similar. Both A and B have a similar gradient.\\
- **Background**: The background should be ignored as it does not affect the texture within the shape.\\
\\
\#\#\# Conclusion:\\
Based on the analysis, **Panel B** contains an identical 2D texture to the shape in panel A. The texture in both panels is a striped pattern with a similar color gradient, despite the different backgrounds and possible rotations. Panels C and D have different textures, with C being a solid color and D having an organic pattern.\\

\includegraphics[width=\linewidth]{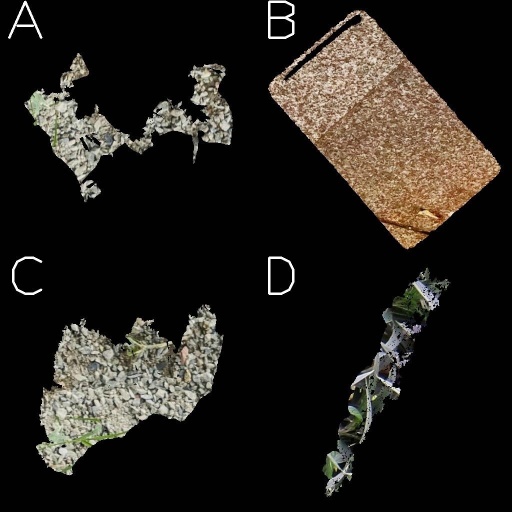} \\
\textbf{Grok 2 response:}\\
To determine which panels contain an identical 2D texture to the shape in panel A, we need to focus on the texture pattern itself, disregarding the shape it appears on, the background, and any potential rotations. Here’s the detailed analysis:\\
1. **Panel A**: The texture in panel A is a rocky, gravel-like pattern with small stones and patches of green vegetation. This is our reference texture.\\
\\
2. **Panel B**: The texture here is a uniform, fine-grained pattern that resembles sandpaper or a textured surface. This texture is distinctly different from the rocky, gravel-like texture in panel A. Therefore, panel B does not contain an identical texture.\\
\\
3. **Panel C**: The texture in panel C is very similar to the texture in panel A. It has the same rocky, gravel-like appearance with small stones and patches of green vegetation. The distribution of the stones and the presence of green patches are consistent with what we see in panel A. Therefore, panel C contains an identical texture.\\
\\
4. **Panel D**: The texture in panel D is different. It has a more leafy, organic pattern with green vegetation that is more prominent and less rocky compared to panel A. The texture is not consistent with the rocky, gravel-like texture of panel A. Therefore, panel D does not contain an identical texture.\\
\\
\#\#\# Conclusion:\\
The panel that contains an identical 2D texture to the shape in panel A is **Panel C**. The distinguishing features used in this analysis were the rocky, gravel-like pattern with small stones and patches of green vegetation, which are present in both panel A and panel C. The other panels either had different textures (B) or different compositions of vegetation and rock (D).

\includegraphics[width=\linewidth]{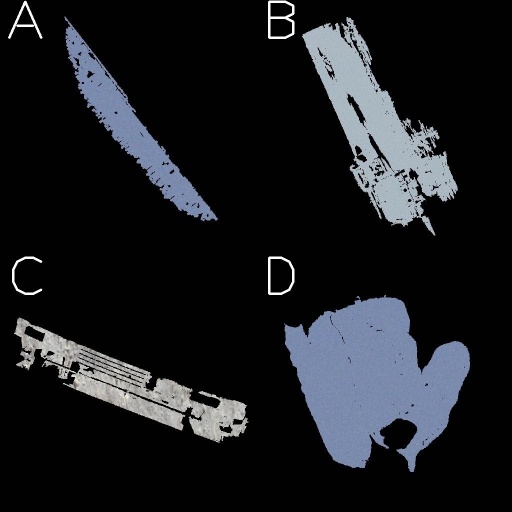} \\
\textbf{Grok 2 response: }\\
To determine which panels contain an identical 2D texture to the shape in panel A, we need to look for distinguishing features in the texture pattern, considering that the texture might be rotated or placed on different shapes with different backgrounds.\\
Here's the detailed analysis:\\
1. **Panel A**:\\
   - The texture in panel A has a specific pattern with irregular shapes and lines. Key features include:\\
 - A prominent diagonal line running through the texture.\\
 - Several smaller, irregularly shaped patches and lines.\\
 - The texture has a blue color with black background.\\
2. **Panel B**:\\
   - The texture in panel B has a similar pattern to panel A, but it is rotated.\\
   - Key features:\\
 - The same diagonal line is visible, though rotated.\\
 - The irregular patches and lines match those in panel A.\\
 - The color and texture pattern are consistent with panel A.\\
3. **Panel C**:\\
   - The texture in panel C is significantly different.\\
   - Key features:\\
 	- The texture appears more uniform and less detailed compared to panel A.\\
 	- The color is lighter and the background is different (white instead of black).\\
 	- The pattern does not match the irregular shapes and lines seen in panel A.\\
4. **Panel D**:\\
   - The texture in panel D is also different.\\
   - Key features:\\
 	- The shape is more rounded and less elongated.\\
 	- The texture pattern lacks the prominent diagonal line and the specific irregular patches seen in panel A.\\
 	- The color and texture pattern are not consistent with panel A.\\
**Conclusion**:\\
The only panel that contains an identical 2D texture to the shape in panel A is **panel B**. This conclusion is based on the matching of the specific texture patterns, including the diagonal line and the irregular patches, despite the rotation and slight differences in the overall shape. Panels C and D do not share the same texture pattern as panel A.

\includegraphics[width=\linewidth]{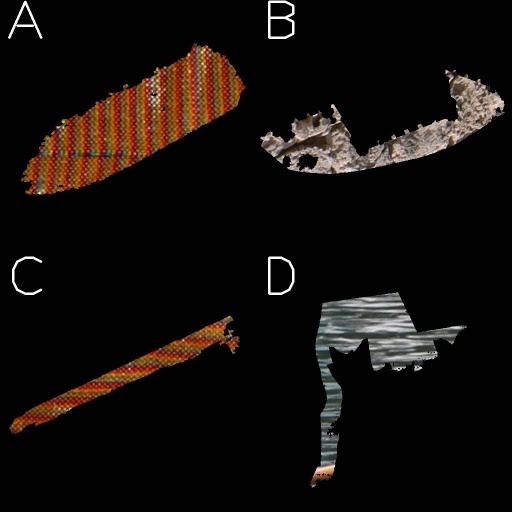} \\
\textbf{Claude 3.5 Sonne response: }\\
Which of the panels contain an identical 2D material/texture  to the shape in panels A?\\
 Note that the shape the texture appear on and the background might be different. The texture might also be rotated.\\
 Explain your answer in detail, including how you got to it and what distinguishing features you used.\\
Looking at the panels, Panel C contains the identical 2D material/texture to Panel A.\\
\\
Here's my detailed analysis:\\
\\
Panel A shows a curved/elongated shape with a distinctive striped pattern in orange, red, and brown colors with a checkered/pixelated texture.\\
\\
Looking at the other panels:\\
- Panel B shows a curved shape with a rough, white/grey textured surface that appears more like a rocky or marble texture\\
- Panel C shows an elongated shape with the same orange-red striped pattern and checkered/pixelated texture as Panel A, just oriented differently\\
- Panel D shows a shape filled with what appears to be a water or metallic ripple texture in grey tones\\
\\
The key distinguishing features I used to match Panel A and C:\\
1. The identical color scheme (orange, red, brown stripes)\\
2. The same pixelated/checkered texture pattern\\
3. The same stripe orientation and spacing, just rotated\\
4. The same level of texture resolution/granularity\\
\\
Therefore, Panel C is the only one that contains the identical texture to Panel A, despite being on a differently shaped outline.\\

\includegraphics[width=\linewidth]{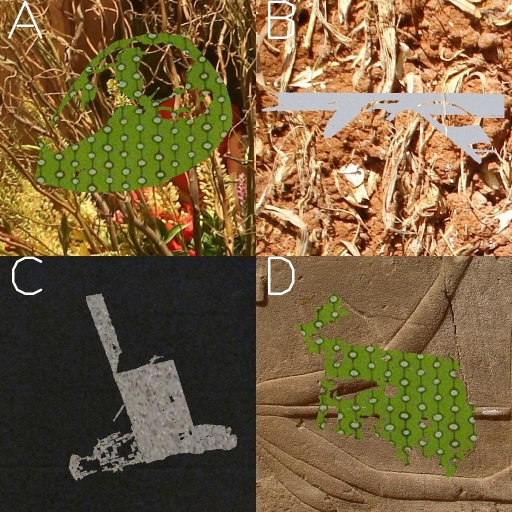} \\
\textbf{Claude 3.5 Sonnet response:  }\\
Looking at all panels, panel D contains the same texture/material as panel A. Here's my detailed analysis:\\
\\
The distinguishing features I used:\\
1. The texture in both A and D shows a green color\\
2. Both have a distinctive polka-dot or circular nodule pattern\\
3. The dots appear to be regularly spaced in a similar grid-like arrangement\\
4. The dots are lighter in color (whitish/grayish) against the green background\\
5. The density and size of the dots appear to be consistent between both panels\\
\\
Panel B and C can be eliminated because:\\
- Panel B shows what appears to be dried plant material or straw on reddish soil\\
- Panel C shows what appears to be a grayscale satellite or aerial image\\
\\
While the texture appears on different shapes (Panel A shows what looks like a bird shape, while Panel D shows what appears to be a different animal shape), and they're against different backgrounds (natural foliage in A, brownish surface in D), the actual texture/material pattern is identical in both panels.\\

\includegraphics[width=\linewidth]{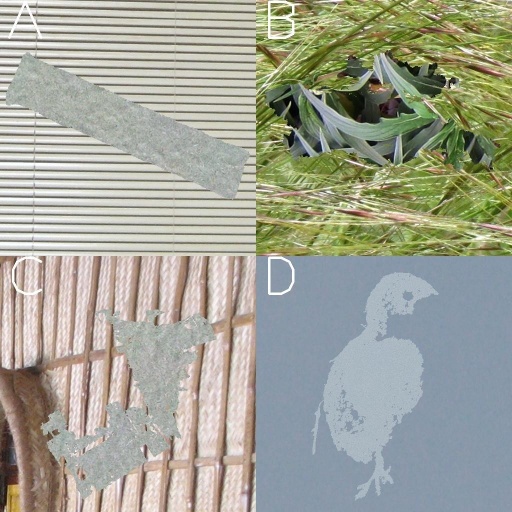} \\
\textbf{Claude 3.5 Sonnet response:  }\\
Which of the panels contain an identical 2D material/texture  to the shape in panels A?\\
 Note that the shape the texture appear on and the background might be different. The texture might also be rotated.\\
 Explain your answer in detail, including how you got to it and what distinguishing features you used.\\
Let's analyze each panel and compare their textures to Panel A:\\
\\
Panel A shows a rectangular strip with a grayish, slightly mottled or granular texture on what appears to be horizontal blinds.\\
\\
Panel B shows what appears to be some green foliage or leaves, with a completely different texture from Panel A.\\
\\
Panel C shows a shape with a grayish, granular texture very similar to Panel A, against what appears to be a corrugated or striped background. The texture appears to have the same mottled, slightly rough appearance as Panel A.\\
\\
Panel D shows a silhouette or shape of what appears to be a bird, with a similar grayish coloring and granular texture to Panel A.\\
\\
Therefore, both Panels C and D contain the same texture/material as Panel A. This can be determined by:\\
1. The similar grayish coloring\\
2. The granular, slightly mottled appearance of the texture\\
3. The way the material catches light and shows subtle variations in its surface\\
While the shapes these textures appear on are different (rectangle in A, irregular shape in C, bird silhouette in D), the underlying material characteristics appear to be identical.\\
The answer is Panels C and D.\\

\section{3D PBR materials: Samples of models responses and explanations. }
\label{appendix:3d_materials_response}
\textbf{Query: Which of the panels contain an object with identical  material/texture to the material of the object   in panels A?}\\
\textbf{ Note that the shape of the object and the setting in which the material appears is different. Explain your answer in detail, including how you got to it and what distinguishing features you used.} \\

\includegraphics[width=\linewidth]{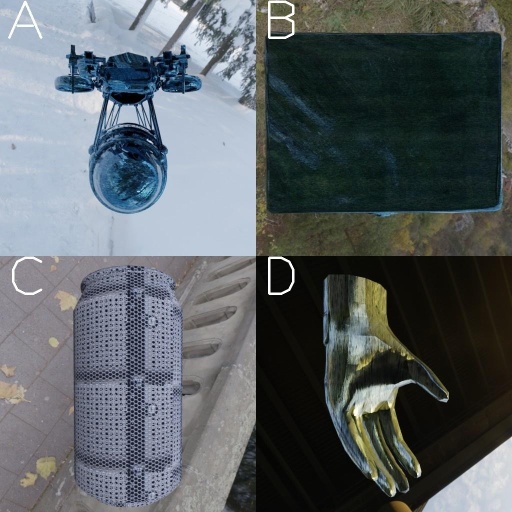} \\
\textbf{GPT 4O Response: } To determine which panel contains an object with an identical material or texture to the object in panel A, let's analyze the images:\\
- **Panel A**: The object has a shiny, metallic, reflective surface. It reflects light and the surroundings, indicating a glossy, mirror-like finish.\\
- **Panel B**: This object has a dark, matte texture, lacking glossiness or reflection. It does not match Panel A’s reflective characteristics.\\
\\
- **Panel C**: The object here appears to have a patterned, mesh-like texture. It does not reflect light similarly to the object in Panel A and does not have the same smooth, reflective surface.\\
\\
- **Panel D**: This object has a reflective, metallic sheen similar to the object in Panel A. The way light reflects off it is consistent with a glossy, metallic surface.\\
\\
Therefore, panel D contains an object with a material and texture similar to the object in panel A. The distinguishing feature used to reach this conclusion was the reflective, glossy surface of the objects in both panels.\\

\includegraphics[width=\linewidth]{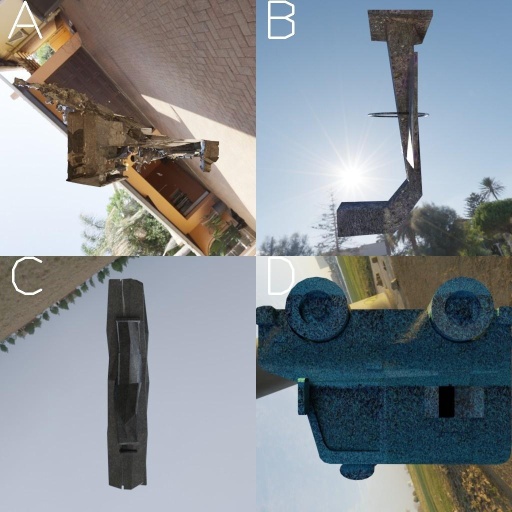} \\
\textbf{GPT 4o  Response:}\\
To identify which panel contains an object with a material/texture identical to that in panel A, we need to closely examine the textures of the objects in each panel.\\
\\
**Panel A:**\\
- The object has a distinct rough, metallic, speckled texture that gives off a shiny, reflective appearance. This texture has varied shades, possibly with a mix of metallic and dark hues.\\
\\
**Panel B:**\\
- The object in this panel appears to have a similar rough texture with reflective qualities. The speckles and irregularities in coloring suggest a material similar to that of Panel A. The similar shift in shades with light also corresponds.\\
\\
**Panel C:**\\
- The object here has a smooth, dark texture, very different from the rough, speckled texture of the object in Panel A. It lacks the metallic appearance.\\
\\
**Panel D:**\\
- This object shows a rough texture but leans more towards a uniform color with a grainy appearance, possibly suggesting a rusted or weathered metallic surface. However, it lacks the reflective, speckled quality of Panel A.\\
\\
**Conclusion:**\\
Panel B contains the object with a material/texture that is identical to the material of the object in Panel A. The key distinguishing features are the rough, speckled texture and the shiny, reflective properties present in both A and B.\\

\includegraphics[width=\linewidth]{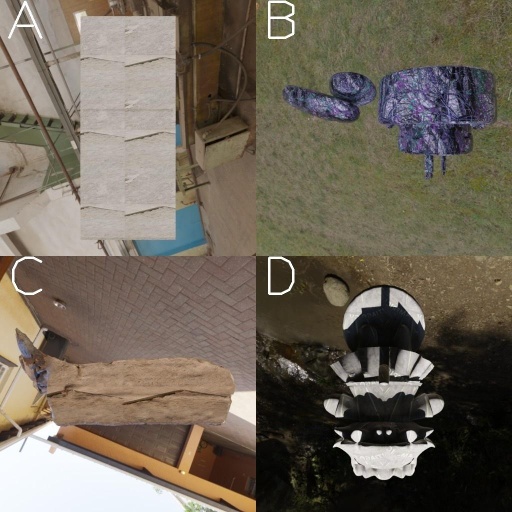} \\
\textbf{GPT 4o Response:}\\
In panel A, the object has a texture that appears to be rough and layered, resembling a form of stone or sedimentary rock.\\
\\
Here's the breakdown for each panel:\\
\\
- **Panel B:** The object has a smooth, shiny surface with a vivid, marbled pattern, which is quite different from the rough, stone-like texture in panel A.\\
\\
- **Panel C:** The object shares a rough and somewhat layered appearance, similar to the stone texture in panel A, making it a match. The object might be oriented differently, but the texture is consistent.\\
\\
- **Panel D:** The object has a smooth, metallic appearance with a reflective surface, differing greatly from the texture in panel A.\\
\\
Given the analysis, panel C contains an object with a texture that matches the material of the object in panel A. The distinguishing feature is the rough, stone-like texture that both objects share.\\

\includegraphics[width=\linewidth]{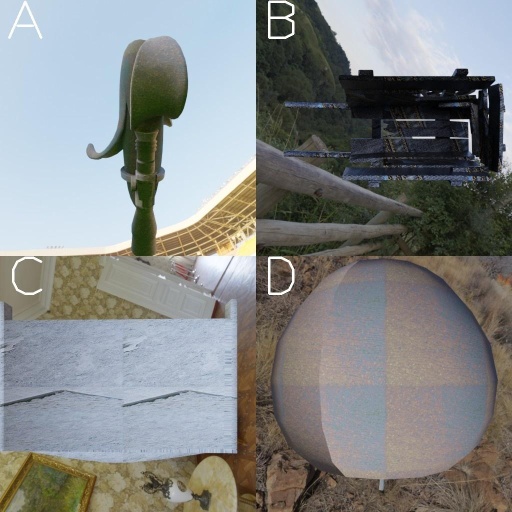} \\
\textbf{GPT 4o Response:}\\
The object in panel A has a green, metallic texture. To identify which other panel contains an object with the same material or texture, let's examine the distinguishing features of the objects in the other panels:\\
\\
1. **Panel B**: The object appears to have a reflective, dark texture that resembles metal but does not match the green color.\\
\\
2. **Panel C**: This panel shows an object with a light, stone-like texture. The color and texture are significantly different from the green metallic look in panel A.\\
\\
3. **Panel D**: The object has a greenish tint and a slightly reflective surface, similar to panel A. It appears that the material in panel D resembles the green metallic texture of panel A, just rotated or with less visible shine due to lighting or surface curvature.\\
\\
Thus, the material in **Panel D** matches the material of the object in panel A. The key features used for comparison were the green tint and the metallic, slightly reflective surface.\\

\includegraphics[width=\linewidth]{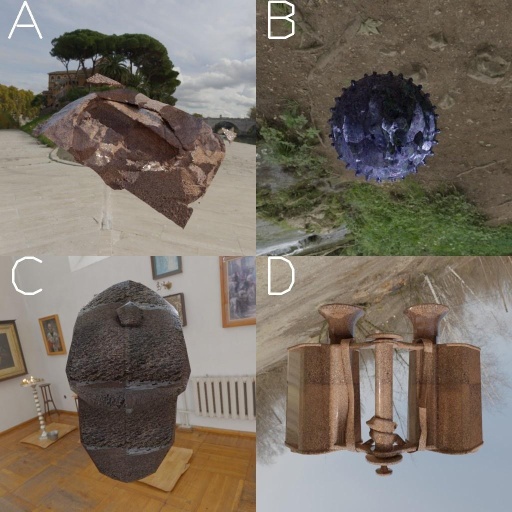} \\
\textbf{Gemini 1.5 pro Response:} Panel C contains an object with a material/texture identical to that in Panel A. Both objects exhibit a rough, layered, and seemingly brittle texture with variations in brown and reddish-brown tones. The material appears to be a type of rock or earthen substance, possibly with some weathering or oxidation. The layered structure and color variations are consistent across both images, suggesting the same material despite the different shapes of the objects.  Panel B shows a smooth, possibly metallic, spiky ball. Panel D presents an object that appears to be made of a more uniform, granular material, possibly wood or a similar composite.  Neither of these share the distinctive layered, rocky texture of Panels A and C.\\

\includegraphics[width=\linewidth]{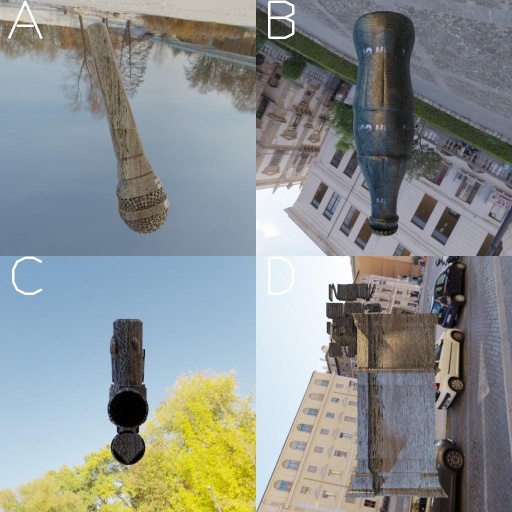} \\
\textbf{Gemini 1.5 pro  Response:}\\
Panels A and C contain objects with the same material/texture.\\
\\
* **Panel A:** Shows what appears to be a microphone made of wood with a distinct bark-like texture and visible growth rings.\\
* **Panel B:** Depicts a soda bottle made of dark, somewhat reflective material – possibly plastic or glass.  It's a smooth, uniform surface with no resemblance to the wood texture.\\
* **Panel C:** Shows an object resembling a small cannon or a decorative element.  The material clearly shows the same bark-like texture and growth rings as the object in Panel A, confirming it's the same type of wood.\\
* **Panel D:** Features a long, rectangular structure made of a weathered, plank-like material. While it has a rougher texture, it lacks the specific bark pattern and rings seen in A and C. It's a different type of wood or possibly a composite material designed to look like wood.\\
The key distinguishing feature is the bark-like texture with growth rings, which is present only in Panels A and C.  This suggests they were created using the same wood material within the 3D modeling or rendering process.\\

\includegraphics[width=\linewidth]{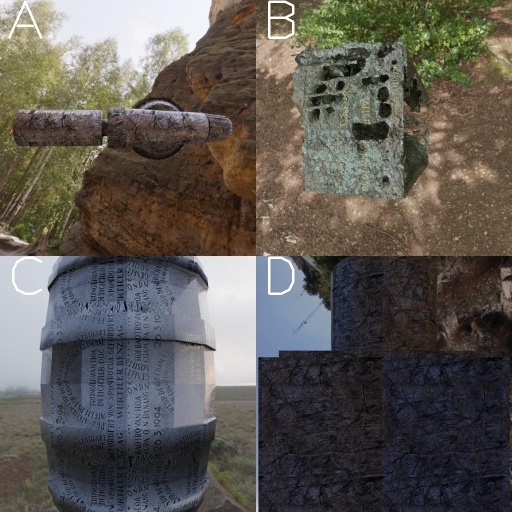} \\
\textbf{Gemini 1.5 pro Response: } Panel D contains an object with the same material/texture as the object in Panel A.\\
The object in Panel A appears to be made of a dark gray material with a rough, cracked texture.  This texture is quite distinctive. Panel D (both the upper and lower portions) showcases the same cracked, rough texture and dark gray coloration. Although Panel D shows the material on a different object (what looks like a section of a pillar or a wall) and in a different environment, the texture itself is a very close match to the object in Panel A.\\
The other panels do not share this texture. Panel B shows an object with a mossy or lichen-covered surface, while Panel C features a smoother surface with writing. Therefore, only Panel D displays the same material properties as Panel A. \\



%% file: main.bib
@article{schwartz1979visual,
  title={Visual shape perception in early infancy},
  author={Schwartz, Marcelle and Day, Ross Henry and Cohen, Leslie B},
  journal={Monographs of the Society for Research in Child Development},
  pages={1--63},
  year={1979},
  publisher={JSTOR}
}

@article{landau1988importance,
  title={The importance of shape in early lexical learning},
  author={Landau, Barbara and Smith, Linda B and Jones, Susan S},
  journal={Cognitive development},
  volume={3},
  number={3},
  pages={299--321},
  year={1988},
  publisher={Elsevier}
}

@article{attneave1956quantitative,
  title={The quantitative study of shape and pattern perception.},
  author={Attneave, Fred and Arnoult, Malcolm D},
  journal={Psychological bulletin},
  volume={53},
  number={6},
  pages={452},
  year={1956},
  publisher={American Psychological Association}
}

@article{imai1994children,
  title={Children's theories of word meaning: The role of shape similarity in early acquisition},
  author={Imai, Mutsumi and Gentner, Dedre and Uchida, Nobuko},
  journal={Cognitive development},
  volume={9},
  number={1},
  pages={45--75},
  year={1994},
  publisher={Elsevier}
}

@article{landy2004visual,
  title={Visual perception of texture},
  author={Landy, Michael S and Graham, Norma and others},
  journal={The visual neurosciences},
  volume={2},
  pages={1106--1118},
  year={2004},
  publisher={Citeseer}
}

@inproceedings{adelson2001seeing,
  title={On seeing stuff: the perception of materials by humans and machines},
  author={Adelson, Edward H},
  booktitle={Human vision and electronic imaging VI},
  volume={4299},
  pages={1--12},
  year={2001},
  organization={spie}
}

@article{julesz1965texture,
  title={Texture and visual perception},
  author={Julesz, Bela},
  journal={Scientific American},
  volume={212},
  number={2},
  pages={38--49},
  year={1965},
  publisher={JSTOR}
}

@article{alayrac2022flamingo,
  title={Flamingo: a visual language model for few-shot learning},
  author={Alayrac, Jean-Baptiste and Donahue, Jeff and Luc, Pauline and Miech, Antoine and Barr, Iain and Hasson, Yana and Lenc, Karel and Mensch, Arthur and Millican, Katherine and Reynolds, Malcolm and others},
  journal={Advances in neural information processing systems},
  volume={35},
  pages={23716--23736},
  year={2022}
}

@inproceedings{li2023blip,
  title={Blip-2: Bootstrapping language-image pre-training with frozen image encoders and large language models},
  author={Li, Junnan and Li, Dongxu and Savarese, Silvio and Hoi, Steven},
  booktitle={International conference on machine learning},
  pages={19730--19742},
  year={2023},
  organization={PMLR}
}

@article{chen2025janus,
  title={Janus-pro: Unified multimodal understanding and generation with data and model scaling},
  author={Chen, Xiaokang and Wu, Zhiyu and Liu, Xingchao and Pan, Zizheng and Liu, Wen and Xie, Zhenda and Yu, Xingkai and Ruan, Chong},
  journal={arXiv preprint arXiv:2501.17811},
  year={2025}
}

@misc{meta2024llama3,
  author       = {Meta AI},
  title        = {Introducing Meta LLaMA 3},
  year         = {2024},
  howpublished = {\url{https://ai.meta.com/blog/meta-llama-3/}},
}

@article{bai2025qwen2,
  title={Qwen2. 5-vl technical report},
  author={Bai, Shuai and Chen, Keqin and Liu, Xuejing and Wang, Jialin and Ge, Wenbin and Song, Sibo and Dang, Kai and Wang, Peng and Wang, Shijie and Tang, Jun and others},
  journal={arXiv preprint arXiv:2502.13923},
  year={2025}
}

@article{achiam2023gpt,
  title={Gpt-4 technical report},
  author={Achiam, Josh and Adler, Steven and Agarwal, Sandhini and Ahmad, Lama and Akkaya, Ilge and Aleman, Florencia Leoni and Almeida, Diogo and Altenschmidt, Janko and Altman, Sam and Anadkat, Shyamal and others},
  journal={arXiv preprint arXiv:2303.08774},
  year={2023}
}

@article{team2024gemini,
  title={Gemini 1.5: Unlocking multimodal understanding across millions of tokens of context},
  author={Team, Gemini and Georgiev, Petko and Lei, Ving Ian and Burnell, Ryan and Bai, Libin and Gulati, Anmol and Tanzer, Garrett and Vincent, Damien and Pan, Zhufeng and Wang, Shibo and others},
  journal={arXiv preprint arXiv:2403.05530},
  year={2024}
}

@inproceedings{fu2024blink,
  title={Blink: Multimodal large language models can see but not perceive},
  author={Fu, Xingyu and Hu, Yushi and Li, Bangzheng and Feng, Yu and Wang, Haoyu and Lin, Xudong and Roth, Dan and Smith, Noah A and Ma, Wei-Chiu and Krishna, Ranjay},
  booktitle={European Conference on Computer Vision},
  pages={148--166},
  year={2024},
  organization={Springer}
}

@article{lu2023mathvista,
  title={Mathvista: Evaluating mathematical reasoning of foundation models in visual contexts},
  author={Lu, Pan and Bansal, Hritik and Xia, Tony and Liu, Jiacheng and Li, Chunyuan and Hajishirzi, Hannaneh and Cheng, Hao and Chang, Kai-Wei and Galley, Michel and Gao, Jianfeng},
  journal={arXiv preprint arXiv:2310.02255},
  year={2023}
}

@article{lee2024vhelm,
  title={Vhelm: A holistic evaluation of vision language models},
  author={Lee, Tony and Tu, Haoqin and Wong, Chi Heem and Zheng, Wenhao and Zhou, Yiyang and Mai, Yifan and Roberts, Josselin and Yasunaga, Michihiro and Yao, Huaxiu and Xie, Cihang and others},
  journal={Advances in Neural Information Processing Systems},
  volume={37},
  pages={140632--140666},
  year={2024}
}

@article{xu2024lvlm,
  title={Lvlm-ehub: A comprehensive evaluation benchmark for large vision-language models},
  author={Xu, Peng and Shao, Wenqi and Zhang, Kaipeng and Gao, Peng and Liu, Shuo and Lei, Meng and Meng, Fanqing and Huang, Siyuan and Qiao, Yu and Luo, Ping},
  journal={IEEE Transactions on Pattern Analysis and Machine Intelligence},
  year={2024},
  publisher={IEEE}
}

@article{rudman2025forgotten,
  title={Forgotten polygons: Multimodal large language models are shape-blind},
  author={Rudman, William and Golovanevsky, Michal and Bar, Amir and Palit, Vedant and LeCun, Yann and Eickhoff, Carsten and Singh, Ritambhara},
  journal={arXiv preprint arXiv:2502.15969},
  year={2025}
}

@article{eppel2024large,
  title={Do large language vision models understand 3D shapes?},
  author={Eppel, Sagi},
  journal={arXiv preprint arXiv:2412.10908},
  year={2024}
}

@article{kamoi2024visonlyqa,
  title={Visonlyqa: Large vision language models still struggle with visual perception of geometric information},
  author={Kamoi, Ryo and Zhang, Yusen and Das, Sarkar Snigdha Sarathi and Zhang, Ranran Haoran and Zhang, Rui},
  journal={arXiv preprint arXiv:2412.00947},
  year={2024}
}

@article{hemmat2024hidden,
  title={Hidden in plain sight: Evaluating abstract shape recognition in vision-language models},
  author={Hemmat, Arshia and Davies, Adam and Lamb, Tom and Yuan, Jianhao and Torr, Philip and Khakzar, Ashkan and Pinto, Francesco},
  journal={Advances in Neural Information Processing Systems},
  volume={37},
  pages={88527--88556},
  year={2024}
}

@inproceedings{leong2024optical,
  title={Optical Illusions Recognition Intelligence},
  author={Leong, Wai Yie and Leong, Yuan Zhi and San Leong, Wai},
  booktitle={2024 IEEE Symposium on Wireless Technology \& Applications (ISWTA)},
  pages={117--120},
  year={2024},
  organization={IEEE}
}

@book{robinson2013psychology,
  title={The psychology of visual illusion},
  author={Robinson, James Outram},
  year={2013},
  publisher={Courier Corporation}
}

@article{adelson1993perceptual,
  title={Perceptual organization and the judgment of brightness},
  author={Adelson, Edward H},
  journal={Science},
  volume={262},
  number={5142},
  pages={2042--2044},
  year={1993},
  publisher={American Association for the Advancement of Science}
}

@book{coren2020seeing,
  title={Seeing is deceiving: The psychology of visual illusions},
  author={Coren, Stanley and Girgus, Joan},
  year={2020},
  publisher={Routledge}
}

@inproceedings{kirillov2023segment,
  title={Segment anything},
  author={Kirillov, Alexander and Mintun, Eric and Ravi, Nikhila and Mao, Hanzi and Rolland, Chloe and Gustafson, Laura and Xiao, Tete and Whitehead, Spencer and Berg, Alexander C and Lo, Wan-Yen and others},
  booktitle={Proceedings of the IEEE/CVF international conference on computer vision},
  pages={4015--4026},
  year={2023}
}

@inproceedings{lin2014microsoft,
  title={Microsoft coco: Common objects in context},
  author={Lin, Tsung-Yi and Maire, Michael and Belongie, Serge and Hays, James and Perona, Pietro and Ramanan, Deva and Doll{\'a}r, Piotr and Zitnick, C Lawrence},
  booktitle={Computer vision--ECCV 2014: 13th European conference, zurich, Switzerland, September 6-12, 2014, proceedings, part v 13},
  pages={740--755},
  year={2014},
  organization={Springer}
}

@inproceedings{raistrick2023infinite,
  title={Infinite photorealistic worlds using procedural generation},
  author={Raistrick, Alexander and Lipson, Lahav and Ma, Zeyu and Mei, Lingjie and Wang, Mingzhe and Zuo, Yiming and Kayan, Karhan and Wen, Hongyu and Han, Beining and Wang, Yihan and others},
  booktitle={Proceedings of the IEEE/CVF conference on computer vision and pattern recognition},
  pages={12630--12641},
  year={2023}
}

@book{nikolenko2021synthetic,
  title={Synthetic data for deep learning},
  author={Nikolenko, Sergey I and others},
  volume={174},
  year={2021},
  publisher={Springer}
}

@article{kuznetsova2020open,
  title={The open images dataset v4: Unified image classification, object detection, and visual relationship detection at scale},
  author={Kuznetsova, Alina and Rom, Hassan and Alldrin, Neil and Uijlings, Jasper and Krasin, Ivan and Pont-Tuset, Jordi and Kamali, Shahab and Popov, Stefan and Malloci, Matteo and Kolesnikov, Alexander and others},
  journal={International journal of computer vision},
  volume={128},
  number={7},
  pages={1956--1981},
  year={2020},
  publisher={Springer}
}

@article{guo2025deepseek,
  title={Deepseek-r1: Incentivizing reasoning capability in llms via reinforcement learning},
  author={Guo, Daya and Yang, Dejian and Zhang, Haowei and Song, Junxiao and Zhang, Ruoyu and Xu, Runxin and Zhu, Qihao and Ma, Shirong and Wang, Peiyi and Bi, Xiao and others},
  journal={arXiv preprint arXiv:2501.12948},
  year={2025}
}

@article{liu2024deepseek,
  title={Deepseek-v3 technical report},
  author={Liu, Aixin and Feng, Bei and Xue, Bing and Wang, Bingxuan and Wu, Bochao and Lu, Chengda and Zhao, Chenggang and Deng, Chengqi and Zhang, Chenyu and Ruan, Chong and others},
  journal={arXiv preprint arXiv:2412.19437},
  year={2024}
}

@article{lu2024deepseek,
  title={Deepseek-vl: towards real-world vision-language understanding},
  author={Lu, Haoyu and Liu, Wen and Zhang, Bo and Wang, Bingxuan and Dong, Kai and Liu, Bo and Sun, Jingxiang and Ren, Tongzheng and Li, Zhuoshu and Yang, Hao and others},
  journal={arXiv preprint arXiv:2403.05525},
  year={2024}
}

@article{li2025benchmark,
  title={Benchmark evaluations, applications, and challenges of large vision language models: A survey},
  author={Li, Zongxia and Wu, Xiyang and Du, Hongyang and Nghiem, Huy and Shi, Guangyao},
  journal={arXiv preprint arXiv:2501.02189},
  volume={1},
  year={2025}
}

@article{zhang2024vision,
  title={Vision-language models for vision tasks: A survey},
  author={Zhang, Jingyi and Huang, Jiaxing and Jin, Sheng and Lu, Shijian},
  journal={IEEE Transactions on Pattern Analysis and Machine Intelligence},
  year={2024},
  publisher={IEEE}
}

@article{schuhmann2021laion,
  title={Laion-400m: Open dataset of clip-filtered 400 million image-text pairs},
  author={Schuhmann, Christoph and Vencu, Richard and Beaumont, Romain and Kaczmarczyk, Robert and Mullis, Clayton and Katta, Aarush and Coombes, Theo and Jitsev, Jenia and Komatsuzaki, Aran},
  journal={arXiv preprint arXiv:2111.02114},
  year={2021}
}

@inproceedings{shilane2004princeton,
  title={The princeton shape benchmark},
  author={Shilane, Philip and Min, Patrick and Kazhdan, Michael and Funkhouser, Thomas},
  booktitle={Proceedings Shape Modeling Applications, 2004.},
  pages={167--178},
  year={2004},
  organization={IEEE}
}

@article{kliot1998invariant,
  title={Invariant-based shape retrieval in pictorial databases},
  author={Kliot, Michael and Rivlin, Ehud},
  journal={Computer Vision and Image Understanding},
  volume={71},
  number={2},
  pages={182--197},
  year={1998},
  publisher={Elsevier}
}

@inproceedings{cao2022unsupervised,
  title={Unsupervised deep multi-shape matching},
  author={Cao, Dongliang and Bernard, Florian},
  booktitle={European Conference on Computer Vision},
  pages={55--71},
  year={2022},
  organization={Springer}
}

@inproceedings{he2018triplet,
  title={Triplet-center loss for multi-view 3d object retrieval},
  author={He, Xinwei and Zhou, Yang and Zhou, Zhichao and Bai, Song and Bai, Xiang},
  booktitle={Proceedings of the IEEE conference on computer vision and pattern recognition},
  pages={1945--1954},
  year={2018}
}

@inproceedings{upchurch2022dense,
  title={A dense material segmentation dataset for indoor and outdoor scene parsing},
  author={Upchurch, Paul and Niu, Ransen},
  booktitle={European conference on computer vision},
  pages={450--466},
  year={2022},
  organization={Springer}
}

@inproceedings{bell2015material,
  title={Material recognition in the wild with the materials in context database},
  author={Bell, Sean and Upchurch, Paul and Snavely, Noah and Bala, Kavita},
  booktitle={Proceedings of the IEEE conference on computer vision and pattern recognition},
  pages={3479--3487},
  year={2015}
}

@inproceedings{drehwald2023one,
  title={One-shot recognition of any material anywhere using contrastive learning with physics-based rendering},
  author={Drehwald, Manuel S and Eppel, Sagi and Li, Jolina and Hao, Han and Aspuru-Guzik, Alan},
  booktitle={Proceedings of the IEEE/CVF International Conference on Computer Vision},
  pages={23524--23533},
  year={2023}
}

@article{deitke2023objaverse,
  title={Objaverse-xl: A universe of 10m+ 3d objects},
  author={Deitke, Matt and Liu, Ruoshi and Wallingford, Matthew and Ngo, Huong and Michel, Oscar and Kusupati, Aditya and Fan, Alan and Laforte, Christian and Voleti, Vikram and Gadre, Samir Yitzhak and others},
  journal={Advances in Neural Information Processing Systems},
  volume={36},
  pages={35799--35813},
  year={2023}
}

@article{lupianiblender,
  title={Blender Scripting with Python},
  author={Lupiani, Isabel},
  publisher={Springer}
}

@article{eppel2024vastextures,
  title={Vastextures: Vast repository of textures and PBR materials extracted from real-world images using unsupervised methods},
  author={Eppel, Sagi},
  journal={arXiv preprint arXiv:2406.17146},
  year={2024}
}

@inproceedings{cimpoi2014describing,
  title={Describing textures in the wild},
  author={Cimpoi, Mircea and Maji, Subhransu and Kokkinos, Iasonas and Mohamed, Sammy and Vedaldi, Andrea},
  booktitle={Proceedings of the IEEE conference on computer vision and pattern recognition},
  pages={3606--3613},
  year={2014}
}

@article{lagunas2019similarity,
  title={A similarity measure for material appearance},
  author={Lagunas, Manuel and Malpica, Sandra and Serrano, Ana and Garces, Elena and Gutierrez, Diego and Masia, Belen},
  journal={arXiv preprint arXiv:1905.01562},
  year={2019}
}

@misc{polyhaven_hdri,
  author       = {{Poly Haven}},
  title        = {Poly Haven HDRI Repository},
  howpublished = {\url{https://polyhaven.com}},
}

@inproceedings{sun2018pix3d,
  title={Pix3d: Dataset and methods for single-image 3d shape modeling},
  author={Sun, Xingyuan and Wu, Jiajun and Zhang, Xiuming and Zhang, Zhoutong and Zhang, Chengkai and Xue, Tianfan and Tenenbaum, Joshua B and Freeman, William T},
  booktitle={Proceedings of the IEEE conference on computer vision and pattern recognition},
  pages={2974--2983},
  year={2018}
}

@article{chollet2024arc,
  title={Arc prize 2024: Technical report},
  author={Chollet, Francois and Knoop, Mike and Kamradt, Gregory and Landers, Bryan},
  journal={arXiv preprint arXiv:2412.04604},
  year={2024}
}

@inproceedings{yang2015large,
  title={A large-scale car dataset for fine-grained categorization and verification},
  author={Yang, Linjie and Luo, Ping and Change Loy, Chen and Tang, Xiaoou},
  booktitle={Proceedings of the IEEE conference on computer vision and pattern recognition},
  pages={3973--3981},
  year={2015}
}

@article{fu2020hard,
  title={Hard example generation by texture synthesis for cross-domain shape similarity learning},
  author={Fu, Huan and Li, Shunming and Jia, Rongfei and Gong, Mingming and Zhao, Binqiang and Tao, Dacheng},
  journal={Advances in neural information processing systems},
  volume={33},
  pages={14675--14687},
  year={2020}
}

@inproceedings{lin2021single,
  title={Single image 3d shape retrieval via cross-modal instance and category contrastive learning},
  author={Lin, Ming-Xian and Yang, Jie and Wang, He and Lai, Yu-Kun and Jia, Rongfei and Zhao, Binqiang and Gao, Lin},
  booktitle={Proceedings of the IEEE/CVF international conference on computer vision},
  pages={11405--11415},
  year={2021}
}

@inproceedings{chen2003visual,
  title={On visual similarity based 3D model retrieval},
  author={Chen, Ding-Yun and Tian, Xiao-Pei and Shen, Yu-Te and Ouhyoung, Ming},
  booktitle={Computer graphics forum},
  volume={22},
  pages={223--232},
  year={2003},
  organization={Wiley Online Library}
}

@article{boyer2011shrec,
  title={SHREC 2011: robust feature detection and description benchmark},
  author={Boyer, Edmond and Bronstein, Alexander M and Bronstein, Michael M and Bustos, Benjamin and Darom, Tal and Horaud, Radu and Hotz, Ingrid and Keller, Yosi and Keustermans, Johannes and Kovnatsky, Artiom and others},
  journal={arXiv preprint arXiv:1102.4258},
  year={2011}
}

@book{chabris2011invisible,
  title={The invisible gorilla: How our intuitions deceive us},
  author={Chabris, Christopher and Simons, Daniel},
  year={2011},
  publisher={Harmony}
}

@article{oquab2023dinov2,
  title={Dinov2: Learning robust visual features without supervision},
  author={Oquab, Maxime and Darcet, Timoth{\'e}e and Moutakanni, Th{\'e}o and Vo, Huy and Szafraniec, Marc and Khalidov, Vasil and Fernandez, Pierre and Haziza, Daniel and Massa, Francisco and El-Nouby, Alaaeldin and others},
  journal={arXiv preprint arXiv:2304.07193},
  year={2023}
}

@article{zhao2022vl,
  title={Vl-checklist: Evaluating pre-trained vision-language models with objects, attributes and relations},
  author={Zhao, Tiancheng and Zhang, Tianqi and Zhu, Mingwei and Shen, Haozhan and Lee, Kyusong and Lu, Xiaopeng and Yin, Jianwei},
  journal={arXiv preprint arXiv:2207.00221},
  year={2022}
}
